\documentclass[pdflatex,sn-mathphys-num]{sn-jnl}
\usepackage{booktabs}

\usepackage{graphicx}%
\usepackage{multirow}%
\usepackage{amsmath,amssymb,amsfonts}%
\usepackage{amsthm}%
\usepackage{mathrsfs}%
\usepackage[title]{appendix}%
\usepackage{xcolor}%
\usepackage{textcomp}%
\usepackage{array}
\usepackage{manyfoot}%
\usepackage{booktabs}%
\usepackage{algorithm}%
\usepackage{algorithmicx}%
\usepackage{algpseudocode}%
\usepackage{listings}%
\usepackage{multirow}
\usepackage{multicol}
\usepackage{booktabs} 

\usepackage{mathtools}

\newcommand{\KL}{\textup{KL}}
\newcommand{\softmax}{\textup{softmax}}
\newcommand{\circled}[1]{\small{\raisebox{.6pt}{\textcircled{\raisebox{-.8pt}{#1}}}}}

\usepackage{tikz} 

\newcounter{optproblem}

\newtheoremstyle{mytheoremstyle} 
    {\topsep}                    
    {\topsep}                    
    {\normalfont}                
    {}                           
    {\bfseries}                   
    {.}                          
    {.5em}                       
    {}  

\theoremstyle{mytheoremstyle}
\newtheorem{theorem}{Theorem}[section]

\newtheorem*{theorem*}{Theorem}
\newtheorem*{lemma*}{Lemma}
\newtheorem*{remark*}{Remark}

\theoremstyle{mytheoremstyle}

\theoremstyle{remark}

\DeclareMathAlphabet{\pazocal}{OMS}{zplm}{m}{n}
\DeclareMathAlphabet{\mathpzc}{OMS}{pzc}{m}{it}

\renewcommand{\hat}{\widehat}

\newcommand{\bfm}[1]{\ensuremath{\mathbf{#1}}}

   \def\bA{\bfm A}  
   \def\bB{\bfm B}  
\def\bc{\bfm c}

  \def\bF{\bfm F}  
     
\def\bh{\bfm h}   \def\bH{\bfm H}  
   \def\bI{\bfm I}  
     
   \def\bK{\bfm K}

   \def\bN{\bfm N}  
     
\def\bp{\bfm p}     
     
     \def\RR{\mathbb{R}}

\def\bu{\bfm u}   \def\bU{\bfm U}  
     
   \def\bW{\bfm W}  
   \def\bX{\bfm X}  
\def\by{\bfm y}   \def\bY{\bfm Y}  
     
\def\bzero{\bfm 0} 

 \def\cA{{\cal  A}}

 \def\cL{{\cal  L}}

 \def\cU{{\cal  U}}
 \def\cV{{\cal  V}}

\def\+#1{\mathcal{#1}}
\def\-#1{\textup{#1}}

\def\set#1{\left\{ #1 \right\}}
\def\pth#1{\left( #1 \right)}
\def\bth#1{\left[ #1 \right]}
\def\abth#1{\left | #1 \right |}

\def\defeq {\coloneqq}

\newcommand{\La}{\left\langle\kern-0.64ex\left\langle}
\newcommand{\Ra}{\right\rangle\kern-0.64ex\right\rangle}

\def\Norm#1#2{{\left\vert\kern-0.4ex\left\vert\kern-0.4ex\left\vert #1
    \right\vert\kern-0.4ex\right\vert\kern-0.4ex\right\vert}_{#2}}
\def\norm#1#2{{\left\|#1\right\|}_{#2}}

\def\lonenorm#1{\norm{#1}{1}}
\def\ltwonorm#1{\norm{#1}{2}}
\def\fnorm#1{\norm{#1}{\textup{F}}}

\newcommand{\rank}{\textup{rank}}

\newcommand{\1}{{\rm 1}\kern-0.25em{\rm I}}
\def\indict#1{{\rm 1}\kern-0.25em{\rm I}_{\set{#1}}}

\def\set#1{\left\{#1\right\}}

\newcommand{\beq}{\begin{equation}}
\newcommand{\eeq}{\end{equation}}
\newcommand{\beqa}{\begin{eqnarray}}
\newcommand{\eeqa}{\end{eqnarray}}
\newcommand{\beqas}{\begin{eqnarray*}}
\newcommand{\eeqas}{\end{eqnarray*}}
\def\bal#1\eal{\begin{align}#1\end{align}}
\def\bals#1\eals{\begin{align*}#1\end{align*}}
\def\bsal#1\esal{\begin{small}\begin{align}#1\end{align}\end{small}}
\def\bsals#1\esals{\begin{small}\begin{align*}#1\end{align*}\end{small}}
\def\bsfal#1\esfal{\begin{small}\begin{flalign}#1\end{flalign}\end{small}}

\theoremstyle{thmstyleone}%

\theoremstyle{thmstyletwo}%

\theoremstyle{thmstylethree}%

\begin{document}

\title[Graph Contrastive Learning with Low-Rank Regularization and Low-Rank Attention for Noisy Node Classification]{Graph Contrastive Learning with Low-Rank Regularization and Low-Rank Attention for Noisy Node Classification}


\author[1]{\fnm{Yancheng} \sur{Wang}}\email{ywan1053@asu.edu}

\author[1]{\fnm{Yingzhen} \sur{Yang}}\email{yingzhen.yang@asu.edu}

\affil[1]{\orgdiv{School of Computing and Augmented Intelligence}, \orgname{Arizona State University}, \orgaddress{\street{699 S Mill Ave}, \city{Tempe}, \postcode{85281}, \state{AZ}, \country{USA}}}


\abstract{Graph Neural Networks (GNNs) have achieved remarkable success in learning node representations and have shown strong performance in tasks such as node classification. However, recent findings indicate that the presence of noise in real-world graph data can substantially impair the effectiveness of GNNs. To address this challenge, we introduce a robust and innovative node representation learning method named Graph Contrastive Learning with Low-Rank Regularization, or GCL-LRR, which follows a two-stage transductive learning framework for node classification. In the first stage, the GCL-LRR encoder is optimized through prototypical contrastive learning while incorporating a low-rank regularization objective. In the second stage, the representations generated by GCL-LRR are employed by a linear transductive classifier to predict the labels of unlabeled nodes within the graph. Our GCL-LRR is inspired by the Low Frequency Property (LFP) of the graph data and its labels, and it is also theoretically motivated by our sharp generalization bound for transductive learning. To the best of our knowledge, our theoretical result is among the first to theoretically demonstrate the advantage of low-rank regularization in transductive learning, which is also supported by strong empirical results. To further enhance the performance of GCL-LRR, we present an improved model named GCL-LR-Attention, which incorporates a novel LR-Attention layer into GCL-LRR. GCL-LR-Attention reduces the kernel complexity of GCL-LRR and contributes to a tighter generalization bound, leading to improved performance. Extensive evaluations on standard benchmark datasets evidence the effectiveness and robustness of both GCL-LRR and GCL-LR-Attention in learning meaningful node representations. 
The code is available at \url{https://github.com/Statistical-Deep-Learning/GCL-LR-Attention}.
}




\maketitle
\section{Introduction}
\label{sec:introduction}
Graph Neural Networks (GNNs) are widely recognized as effective tools for node representation learning, with numerous models~\citep{kipf2017semi,bruna2013spectral,hamilton2017inductive,xu2018powerful} demonstrating success across a range of tasks. Among them, methods such as~\citep{kipf2017semi, zhu2020simple} utilize the graph topology by aggregating information from neighboring nodes. This form of feature propagation enables the models to generate powerful embeddings, leading to strong outcomes in semi-supervised node classification and clustering.
However, the majority of existing GNN methods do not adequately address the presence of noise in the graph data~\citep{zhu2024robust, zhong2019graph}. Such noise may arise in the attributes or labels of nodes, introducing attribute noise and label noise, respectively. Prior studies~\citep{patrini2017making} have demonstrated that noise in the input data can significantly impair the generalization ability of neural networks. In the case of graphs, this issue is further amplified since noise associated with a few nodes can spread through the graph structure and affect other nodes~\citep{dai2021nrgnn, wang2023deep, WangSZWFHB24}. As a result, corrupted nodes not only degrade their own representations but also influence those of their neighbors. Although it is possible to manually clean or relabel data, such approaches are costly and do not scale well. This challenge underscores the need for GNN models that are capable of learning effectively even in the presence of noisy inputs.

To this end, we introduce Graph Contrastive Learning with Low-Rank Regularization, abbreviated as GCL-LRR, which is a new encoder aimed at improving both robustness and generalization in node representation learning. Traditional strategies for robust learning generally fall into two categories. The first modifies the loss function to accommodate corrupted data~\citep{patrini2017making,goldberger2016training}, while the second removes samples suspected to be noisy~\citep{malach2017decoupling,jiang2018mentornet,yu2019does,li2020dividemix,Han2018NIPS}. Although some of these ideas have been adapted for graph settings~\citep{dai2021nrgnn, qian2022robust, zhuang2022defending}, they often depend on heuristics and lack theoretical backing in semi-supervised scenarios.
The design of our GCL-LRR model is inspired by the low-frequency nature of graph signals, and it is also theoretically supported by a new generalization bound developed for transductive learning. To the best of our knowledge, this paper is among the first to provide a principled justification for the advantage of low-rank representation learning in graph contrastive settings. Experimental evaluations conducted on widely used benchmarks demonstrate that GCL-LRR consistently achieves strong robustness and high performance.
Although GNNs are known to function as low-pass filters, they do not explicitly target low-frequency signals. Consequently, their ability to exploit the Low-Frequency Property (LFP) in noisy labels remains limited. As visualized in Figure~\ref{fig:eigen-projection}, the LFP reveals that clean label information tends to concentrate within the low-rank structure of the observed label space. Unlike conventional GNNs, the GCL-LRR encoder explicitly captures this structure by learning representations constrained to be low-rank. Prior work~\citep{cheng2021graph} has illustrated the benefit of such low-rank learning in mitigating attribute noise by introducing learnable filtering mechanisms. Moreover, recent studies involving graph attention and transformer-based architectures emphasize the need to balance both low- and high-frequency components for improved node representations~\citep{GFSA, zhang2024hongat}. In comparison, GCL-LRR offers an improved balance by promoting low-frequency information through minimization of the Truncated Nuclear Norm (TNN), which aligns with the LFP.
Our GCL-LRR method is also grounded in theory through our novel generalization bound in the transductive setting. To further boost the performance of GCL-LRR, we introduce an enhanced model termed GCL-LR-Attention, which integrates a novel LR-Attention layer into GCL-LRR. GCL-LR-Attention lowers the kernel complexity of GCL-LRR by the LR-Attention layer and yields a tighter generalization error bound than GCL-LRR, leading to even better performance than GCL-LRR. Performance comparisons in Table~\ref{table:label_noise_short} of Section~\ref{sec:classification} demonstrate that GCL-LRR and GCL-LR-Attention surpass state-of-the-art methods based on attention and transformers, such as GFSA~\citep{GFSA} and HONGAT~\citep{zhang2024hongat}, when evaluated under label and attribute noise. Furthermore, the trade-off between low- and high-frequency learning is evaluated through kernel complexity, a metric detailed in Section~\ref{sec:low-rank-transductive}. As shown in Table~\ref{tab:upper_bound} of Section~\ref{sec:kernel_complexity}, the GCL-LRR model achieves lower kernel complexity, leading to a lower upper bound on the test loss than competing graph contrastive and attention-based methods, confirming the effectiveness of the learned low-rank representations. Benefiting from the LR-Attention layer, GCL-LR-Attention achieves an even lower generalization bound and kernel complexity, leading to even better node classification performance under various types of noise.

\subsection{Contributions}
Our contributions are as follows.

First, we introduce Graph Contrastive Learning with Low-Rank Regularization (GCL-LRR), a novel and theoretically grounded GCL encoder. The design of GCL-LRR is motivated by the LFP illustrated in Figure~\ref{fig:eigen-projection}, which shows that a low-rank projection of the clean label matrix captures most of its informative content. In contrast, label noise tends to spread uniformly across all eigenvectors of the classification kernel matrix. Drawing from this insight, GCL-LRR incorporates the TNN as a low-rank regularization component into the loss function of standard prototypical graph contrastive learning.
This regularization encourages the model to produce inherently low-rank representations, which are subsequently used in a linear transductive classifier.

Second, we establish a strong theoretical foundation that guarantees the generalization performance of the linear transductive algorithm when applied to the low-rank features generated by GCL-LRR. In particular,
we derive a novel generalization bound on the test loss for unlabeled nodes. To the best of our knowledge, this theoretical result is among the first demonstrating the benefit of learning low-rank representations for robust transductive classification under label noise.
Motivated by this theoretical result and in pursuit of further performance gains, we introduce an enhanced variant of our model that incorporates a novel LR-Attention layer, termed the GCL-LR-Attention. GCL-LR-Attention achieves a further reduction in kernel complexity compared to GCL-LRR, which leads to a tighter bound on the generalization error in the transductive setting and improved empirical performance.
As demonstrated in Table~\ref{tab:upper_bound} of Section~\ref{sec:kernel_complexity}, the GCL-LRR model has a lower generalization upper bound than existing methods. The GCL-LR-Attention model further reduces the generalization upper bound of the GCL-LRR model.
Comprehensive experiments conducted on widely used graph benchmarks demonstrate the superiority of both GCL-LRR and GCL-LR-Attention over existing methods in node classification tasks involving noisy graph data.

\section{Related Works}
\label{sec:related_works}
\subsection{Graph Neural Network and its Training on Noisy Data}
Graph neural networks (GNNs) have established themselves as effective approaches for learning node representations. Existing GNNs are generally grouped into spectral methods~\citep{bruna2013spectral,kipf2017semi} and spatial methods~\citep{hamilton2017inductive,velivckovic2017graph,xu2018powerful}, both of which learn by aggregating information from the local neighborhoods of nodes. In recent years, contrastive learning has become increasingly prominent for unsupervised learning on graphs~\citep{suresh2021adversarial,thakoor2021bootstrapped,wang2022augmentation,lee2022augmentation,feng2022adversarial,zhang2023spectral,lin2023spectral}. A majority of graph contrastive learning (GCL) methods~\citep{velickovic_2019_iclr,sun2019infograph,hu2019strategies,jiao2020sub,peng2020graph,you2021graph,Jin2021MultiScaleCS,mo2022simple} generate multiple augmented views of the original graph and encourage the alignment between the embeddings corresponding to each view. In addition to augmenting at the node level, some recent efforts~\citep{xu2021self,guo2022hcsc,PCL} have improved GCL by introducing semantic prototypes~\citep{snell2017prototypical,arik2020protoattend,allen2019infinite,xu2020attribute} and using these prototypes to contrast node features against global representations.

A major challenge in training deep neural networks, including GNNs, lies in their vulnerability to noisy inputs, which has been highlighted in earlier research~\citep{zhang2021understanding}. Work on robust learning generally follows two principal strategies. One approach adjusts the objective function during training to account for corrupted data, a technique commonly known as loss correction~\citep{patrini2017making,goldberger2016training}. Another approach focuses on identifying clean samples from noisy datasets and using them for training, a process referred to as sample selection~\citep{malach2017decoupling,jiang2018mentornet,yu2019does,li2020dividemix,Han2018NIPS}.
Within the domain of graph-based learning, recent contributions aim to improve robustness by incorporating techniques such as structural prediction, denoising of labels, and self-supervised objectives~\citep{dai2021nrgnn,qian2022robust,zhuang2022defending, CRGNN, CGNN}. Departing from these earlier directions, our approach focuses on enhancing the robustness of GNN encoders for node classification by embedding low-rank regularization directly into the training process of the GCL encoder.
\subsection{Learning Low-Frequency Signal in Graphs with GNNs and Graph Attention}

Traditional graph neural networks, such as GCN~\citep{kipf2017semi}, operate by aggregating information from neighboring nodes and therefore naturally function as low-pass filters. Earlier investigations~\citep{Hoang2019revisit, XuSCCC19, WuSZFYW19, yu2020graph} underscore the significance of capturing low-frequency signals that are inherent in both graph structures and node features. Nevertheless, an exclusive reliance on low-frequency components can introduce the problem of over-smoothing~\citep{bo2021beyond, zhang2024beyond, dong2025graph, sun2022improving}, which becomes particularly detrimental in scenarios where inter-class links are abundant. To address this issue, recent studies~\citep{bo2021beyond, dong2021adagnn, ju2022adaptive} have introduced adaptive techniques that aim to balance low- and high-frequency signals, achieving better results in node classification tasks~\citep{tang2025fahc}. Alongside these developments, models that prioritize learning from low-rank components of graph features and structures have been shown to be more robust in the presence of noise~\citep{TangNAA24, yang2023graph}. Graph attention mechanisms such as GAT~\citep{velivckovic2017graph} are also capable of emphasizing low-frequency patterns~\citep{zhang2024hongat, W0SLTP24}. For instance, HONGAT~\citep{zhang2024hongat} combats the over-smoothing effect by enhancing high-order dependencies and introducing sparsity into the attention weights. Additionally, combining spectral filtering with attention mechanisms has emerged as a promising direction to enable adaptive frequency-aware learning in node representation models~\citep{chang2021spectral, sun2024high, wang2024attention}.

\section{Problem Setup}
\label{sec:setup}
\subsection{Notations}
Let $\mathcal{G} = (\mathcal{V}, \mathcal{E}, \bX)$ denote an attributed graph with $N$ nodes. The node set is given by $\mathcal{V} = \{v_1, v_2, \dots, v_N\}$ and the edge set satisfies $\mathcal{E} \subseteq \mathcal{V} \times \mathcal{V}$. The matrix $\bX \in \mathbb R^{N \times D}$ contains the attribute information of all nodes, where $D$ corresponds to the dimensionality of each node’s attributes. The adjacency matrix associated with $\mathcal{G}$ is denoted by $\mathbf{A} \in \{0, 1\}^{N \times N}$ and satisfies $\mathbf{A}_{ij} = 1$ whenever there is an edge $(v_i, v_j)$ in $\mathcal{E}$. When self-loops are added to the graph, the resulting adjacency matrix becomes $\Tilde{\mathbf{A}} = \mathbf{A} + \mathbf{I}$, and the corresponding degree matrix is defined as $\Tilde{\mathbf{D}}$, which is diagonal. The notation $[N]$ refers to the set of integers from $1$ to $N$ inclusive. A subset $\cL \subseteq [N]$ contains $m$ labeled nodes, and its complement $\cU = [N] \setminus \cL$ has cardinality $u$. The sets $\cV_{\cL}$ and $\cV_{\cU}$ represent the collections of labeled and unlabeled nodes respectively, with $\abth{\cV_{\cL}} = m$ and $\abth{\cV_{\cU}} = u$. For any vector $\bu \in \RR^N$, the notation $\bth{\bu}_{\cA}$ refers to the subvector composed of entries indexed by $\cA \subseteq [N]$. In the case where $\bu$ is a matrix, $\bth{\bu}_{\cA}$ denotes the submatrix consisting of the rows indexed by $\cA$. The Frobenius norm of a matrix is denoted by $\fnorm{\cdot}$, and the $p$-norm of a vector is expressed as $\norm{\cdot}{p}$.
\subsection{Graph Convolution Network}
\label{sec:GCN-backbone}
A straightforward yet effective approach for learning node representations from the attribute matrix $\bX$ and the graph structure $\mathbf{A}$ is the Graph Convolutional Network (GCN). Originally introduced for semi-supervised node classification, GCN is composed of two layers of graph convolution operations. In our framework, we adopt GCN as the base architecture within the proposed GCL-LRR, which serves as the GCL encoder, to generate node embeddings denoted by $\hat\bH \in \RR^{N\times d}$. Each row $\hat\bH_i$ corresponds to the representation of node $v_i$. The computation in GCL-LRR follows the formulation $\hat\bH = g(\bX,\mathbf{A}) = \sigma(\hat{\mathbf{A}}\sigma(\hat{\mathbf{A}}\bX{\tilde \bW}^{(0)}){\tilde \bW}^{(1)})$, where the matrix $\hat{\mathbf{A}}$ is defined as ${\Tilde{\mathbf{D}}}^{-1/2}\Tilde{\mathbf{A}}\Tilde{\mathbf{D}}^{-1/2}$, and the parameters ${\tilde \bW}^{(0)}$ and ${\tilde \bW}^{(1)}$ are learnable weight matrices. The non-linear transformation $\sigma$ represents the ReLU activation function. The resulting node representations produced by GCL-LRR are both robust and low-rank, and they are subsequently used as input to a linear classifier for transductive node classification. Further details regarding the GCL-LRR encoder and the linear transductive classification procedure are provided in Section~\ref{sec:formulation}.

\subsection{Problem Description}
\label{sec:prob-description}
In real-world graph data, noise frequently appears either in the node attributes or in the labels. Such noise can significantly deteriorate the quality of node representations produced by standard GCL encoders, which in turn impairs the performance of classifiers trained on these representations. Our objective is to develop node embeddings that remain robust under two specific scenarios. The first involves noise present in the labels of $\mathcal{V_L}$, while the second concerns noise within the input node attributes $\bX$. These two cases correspond to noisy labels and noisy features, respectively.

The GCL-LRR model is designed to learn node representations that are low-rank and resilient to these types of noise. The representations are defined as $\bH = g(\bX,\mathbf{A})$, where $g(\cdot)$ denotes the GCL-LRR encoder. In this work, the encoder $g$ is implemented as a two-layer GCN, as described in Section~\ref{sec:GCN-backbone}. The output $\bH = \set{\bh_1;\bh_2;\ldots;\bh_N} \in \RR^{N \times d}$ consists of low-rank embeddings for all nodes. These representations are then used for transductive node classification. During this process, a linear classifier is trained using the labeled node set $\mathcal{V_L}$, and the trained classifier is subsequently applied to predict labels for the unlabeled test nodes in $\mathcal{V_U}$.


\section{Methods}
\label{sec:formulation}
\subsection{Low-Rank GCL: Graph Contrastive Learning with Low-Rank Regularization}
\textbf{Preliminary of Prototypical GCL.} Node representation learning focuses on training an encoder $g(\cdot)$, which is implemented using a two-layer Graph Convolutional Network (GCN)~\citep{kipf2017semi}, to generate embeddings that are discriminative and informative. In this work, we adopt a contrastive learning framework to optimize the GCL encoder $g(\cdot)$. Two augmented graph views, denoted as $G^{1} = (\bX^{1}, \mathbf{A}^{1})$ and $G^{2} = (\bX^{2}, \mathbf{A}^{2})$, are created by applying transformations such as node dropping, edge perturbation, and attribute masking. The resulting node representations are given by $\bH^{1}=g(\bX^{1}, \mathbf{A}^{1})$ and $\bH^{2}=g(\bX^{2}, \mathbf{A}^{2})$, where $\bH^1_i$ and $\bH^2_i$ refer to the representations of the $i$-th node in the respective views.

The objective is to enhance the mutual information between $\bH^{1}$ and $\bH^{2}$, and this is achieved by optimizing a lower bound on mutual information through the InfoNCE loss~\citep{PCL}, which serves as the node-level contrastive objective. In order to encode more semantic structure in the learned representations, we also incorporate prototypical contrastive learning~\citep{PCL}, which encourages alignment between node embeddings and a set of cluster prototypes $\set{\bc_1,...,\bc_{{K}}}$. Following the approaches described in~\citep{PCL, snell2017prototypical}, we perform $K$-means clustering on the set $\set{\bh_i}_{i=1}^{N}$ to obtain the $K$ clusters $\set{S_k}_{k=1}^{K}$. Each cluster prototype $\bc_k$ is computed as $\bc_k =\frac{1}{|S_k|} \sum_{\bh_i\in S_k} \bh_i$ for every $k$ in $[K]$. The overall objective for Prototypical GCL consists of two components, $\mathcal{L}_{\textup{node}}$, which corresponds to node-level contrastive loss, and $\mathcal{L}_{\textup{proto}}$, which accounts for the prototype-based contrastive learning. $\mathcal{L}_{\textup{node}}$ and $\mathcal{L}_{\textup{proto}}$ are computed as
\bals
\mathcal{L}_{\textup{node}} &= -\frac{1}{N}\sum_{i=1}^N  \log \frac{s(\bH^1_i, \bH^2_i)}{s(\bH^1_i, \bH^2_i)+ \sum_{j=1}^{N} s(\bH^1_i,\bH^2_j) },  \\
\mathcal{L}_{\textup{proto}} &= -\frac{1}{N}\sum_{i=1}^N\log\frac{\exp(\bH_i\cdot \mathbf{c}_{k}/\tau)}{\sum_{k=1}^{{K}} \exp(\bH_i \cdot \mathbf{c}_k/\tau)},
\eals
where $s(\bH^1_i, \bH^2_i) $ is the cosine similarity between $\bH^1_i$ and $\bH^2_i$.


\noindent\textbf{GCL-LRR: Graph Contrastive Learning with Low-Rank Regularization.} GCL-LRR enhances the robustness and generalization of node representations derived from Prototypical GCL by promoting a low-rank learned feature kernel. The gram matrix $\bK$ of the node representations $\bH \in \RR^{N \times d}$ is calculated by $\bK = \bH^{\top}\bH \in \RR^{N \times N}$.  Let $\set{\hat \lambda_i}_{i=1}^{N}$ with $\hat \lambda_1 \ge \hat \lambda_2 \ldots \ge \hat \lambda_{\min\set{N,d}} \ge \hat \lambda_{\min\set{N,d}+1} = \ldots, = 0$ be the eigenvalues of $\bK$. In order to encourage the features $\bH$ or the gram matrix $\bH^{\top}\bH$ to be low-rank, we explicitly add the TNN $ \norm{\bK}{r_0} \defeq \sum_{r=r_0+1}^{N} \hat \lambda_i$ to the loss function of prototypical GCL. The starting rank $r_0<\min(N,d)$ is the rank of the gram matrix of the features we aim to obtain with the GCL-LRR encoder, that is, if $\norm{\bK}{r_0} = 0$, then $\rank(\bK) = r_0$. Therefore, the overall loss function of GCL-LRR is
\bal
\label{eq:loss-GCL-LR-overall}
\mathcal{L}_{\textup{GCL-LRR}} = \mathcal{L}_{\textup{node}} + \mathcal{L}_{\textup{proto}}  + \tau \norm{\bK}{r_0},
\eal
where $\tau > 0$ is the weighting parameter for the TNN $ \norm{\bK}{r_0}$. We summarize the training algorithm for the GCL-LRR encoder in Algorithm~\ref{Algorithm-RGCL}. After the training is completed, we compute the low-rank node feature by $\bH = g(\bA,\bX)$.
Algorithm~\ref{Algorithm-RGCL} outlines the training procedure for Graph Contrastive Learning with Low-Rank Regularization (GCL-LRR). The characteristics of the datasets employed in our experimental evaluations are summarized in Table~\ref{tab:dataset}.

\begin{algorithm}[tbp]
\caption{Graph Contrastive Learning with Low-Rank Regularization (GCL-LRR)}
\label{Algorithm-RGCL}
\begin{algorithmic}[1]
\Require Attribute matrix $\bX$, adjacency matrix $\mathbf{A}$, training epochs $t_{\text{max}}$
\Ensure Parameters of GCL-LRR encoder $g$
\State Initialize parameters of encoder $g$
\For{$t = 1$ to $t_{\text{max}}$}
    \State Compute node representations: $\bH = g(\bX, \mathbf{A})$
    \State Generate augmented views $G^{1}, G^{2}$ with $\bX^{1}, \mathbf{A}^{1}$ and $\bX^{2}, \mathbf{A}^{2}$
    \State Compute $\bH^{1} = g(\bX^{1}, \mathbf{A}^{1})$, $\bH^{2} = g(\bX^{2}, \mathbf{A}^{2})$
    \State Cluster $\{\bh_i\}_{i=1}^{N}$ into $K$ clusters $\{S_k\}_{k=1}^{K}$ using $K$-means
    \For{$k = 1$ to $K$}
        \State Compute prototype: $\bc_k = \frac{1}{|S_k|} \sum_{\bh_i \in S_k} \bh_i$
    \EndFor
    \State Compute eigenvalues $\{\lambda_i\}_{i=1}^{N}$ of kernel matrix $\bK=\bH^\top \bH$
    \State Update parameters of $g$ using gradient descent on $\mathcal{L}_{\textup{GCL-LRR}}$
\EndFor
\State \Return Encoder $g$
\end{algorithmic}
\end{algorithm}

\begin{figure*}[!htb]
\centering
\includegraphics[width=1\textwidth]{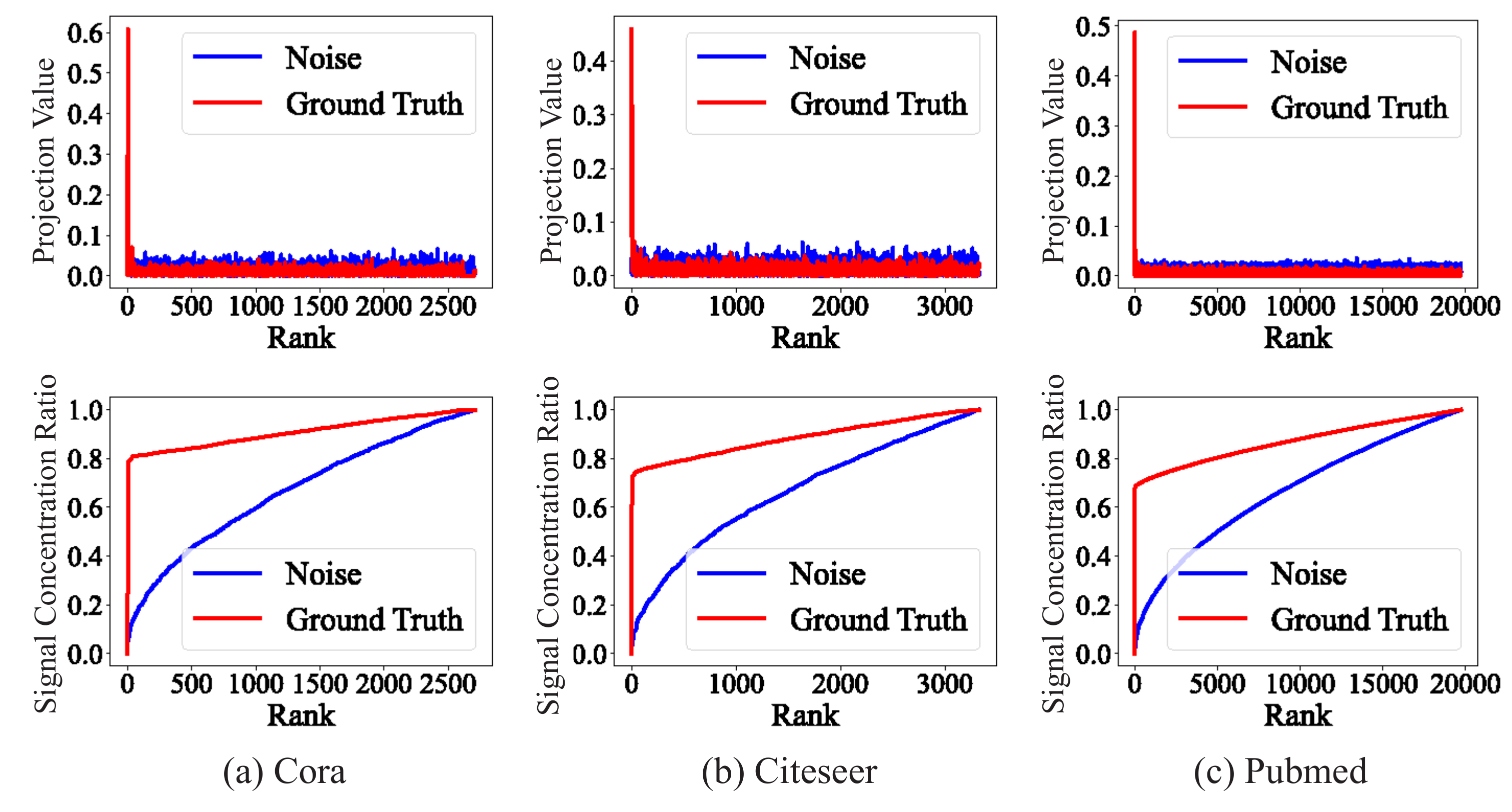}
\caption{Eigen-projection (first row) and signal concentration ratio (second row) on Cora, Citeseer, and Pubmed. The study in this figure is performed for asymmetric label noise with a noise level of $60\%$. By the rank $r=0.2\min\set{N,d}$, the signal concentration ratio of $\tilde \bY$ for Cora, Citeseer, and Pubmed are $0.844$, $0.809$, and $0.784$ respectively. Figure~\ref{fig:eigen-projection-more} in Section~\ref{sec:sup_projection} further illustrates the eigen-projection and signal concentration ratio on more datasets.}
\label{fig:eigen-projection}
\end{figure*}

\noindent\textbf{Motivation of Learning Low-Rank Features.}
We investigate how the information from the ground-truth clean labels and the label noise is distributed across different eigenvectors of the feature gram matrix $\bK$ through an eigen-projection analysis. Let $\tilde \bY \in \RR^{N \times C}$ denote the clean label matrix without noise. We begin by computing the eigenvectors $\bU$ of the gram matrix $\bK = \bH \bH^{\top}$. The eigen-projection score for the $r$-th eigenvector is then given by $p_r = \frac{1}{C}\sum_{c=1}^{C} \ltwonorm{{\bU^{(r)}}^{\top} \tilde \bY^{(c)}}^2/ \ltwonorm{\tilde \bY^{(c)}}^2$ for $r \in [N]$, where $C$ is the number of classes, and $\tilde \bY \in {0,1}^{N \times C}$ consists of one-hot encoded clean labels. Here $\tilde \bY^{(c)}$ refers to the $c$-th column of $\tilde \bY$. We define $\bp = \bth{p_1,\ldots,p_N} \in \RR^N$ as the vector of projection values. In the presence of label noise $\bN \in \RR^{N \times C}$, the observed label matrix becomes $\bY = \tilde \bY + \bN$. The projection value $p_r$ quantifies the proportion of the signal aligned with the $r$-th eigenvector of $\bK$, while the signal concentration ratio at rank $r$ of the ground truth class label is defined as $\lonenorm{\bp^{(1:r)}}$, representing the cumulative contribution of the top $r$ eigenvectors.
Similarly, the signal concentration ratio at rank $r$ of the noise is defined as $\frac{1}{C}\sum_{c=1}^{C} \ltwonorm{{\bU^{(r)}}^{\top}\bN^{(c)}}^2/ \ltwonorm{\bN^{(c)}}^2 \in \RR^N$.
Empirical results, shown as red curves in Figure~\ref{fig:eigen-projection}, indicate that the clean label signals are primarily concentrated on the leading eigenvectors of $\bK$. In contrast, the projection of the label noise appears more uniformly distributed across all eigenvectors, as demonstrated by the blue curves in the same figure.
The above observation motivates low-rank features $\bH$ or, equivalently, the low-rank gram matrix $\bK$. This is because the low-rank part of the feature matrix $\bH$ or the gram matrix $\bK$ covers the dominant information in the ground truth label $\tilde \bY$ while learning only a small portion of the label noise.
We refer to such property as the Low Frequency Property (\textbf{LFP}), which has been widely studied in deep learning~\citep{rahaman19a-spectral-bias,AroraDHLW19-fine-grained-two-layer,CaoFWZG21-spectral-bias,ChorariaD0MC22-spectral-bias-pnns, WangGNSWY24}.
Moreover, we remark that the regularization term $\norm{\bK}{r_0}$ in the loss function (\ref{eq:loss-GCL-LR-overall}) of GCL-LRR is also theoretically motivated by the sharp upper bound for the test loss using a linear transductive classifier, presented as (\ref{eq:optimization-linear-kernel-test-loss}) in Theorem~\ref{theorem:optimization-linear-kernel}. A smaller $\norm{\bK}{r_0}$ renders a smaller upper bound for the test loss, which ensures better generalization capability of the linear transductive classifier to be introduced in the next subsection.

\subsection{Transductive Node Classification}
\label{sec:low-rank-transductive}
In this section, we present a standard yet straightforward linear transductive node classification method based on the low-rank node representations $\bH \in \RR^{N \times d}$ obtained from the GCL-LRR encoder. Furthermore, we establish a novel generalization bound on the test loss of this transductive classifier in the presence of label noise.

We begin by introducing the fundamental notations for the proposed algorithm. For each node $v_i$ where $i \in [N]$, let $\by_i \in \RR^C$ denote its observed one-hot class label vector. We then define the full observed label matrix as $\bY \defeq \bth{\by_1;\by_2;\ldots;\by_N} \in \RR^{N \times C}$, which may include label noise, modeled as $\bN \in \RR^{N \times C}$. The classifier's linear prediction is defined by $\bF(\bW) = \bH \bW$, where $\bW \in \RR^{d \times C}$ is the learnable weight matrix. Final predictions are made using the softmax transformation $\softmax(\bF(\bW)) \in \RR^{N \times C}$ to estimate class probabilities for the test nodes. We train the transductive classifier by minimizing the regular cross-entropy on the labeled nodes through the following optimization problem,
\bal\label{eq:node-classification-obj}
\min\limits_{\bW} L(\bW) =  \frac{1}{m} \sum_{v_i \in \mathcal{V_L}} \KL \pth{\by_i, \bth{\softmax\pth{\bH \bW}}_i},
\eal
where $\KL$ is the KL divergence between the label $\by_i$ and the softmax of the classifier output at node $v_i$. We use a regular gradient descent to optimize (\ref{eq:node-classification-obj}) with a learning rate $\eta \in (0,\frac 1{\hat \lambda_1})$. $\bW$ is initialized by $\bW^{(0)} = \bzero$, and at the $t$-th iteration of gradient descent for $t \ge 1$, $\bW$ is updated by $\bW^{(t)} =  \bW^{(t-1)} - \eta \nabla_{\bW} L (\bW) \vert_{\bW = \bW^{(t-1)}}$.
We define $\bF(\bW,t) \defeq \bH \bW^{(t)}$ as the output of the classifier after the $t$-th iteration of gradient descent for $t \ge 1$. We have the following theoretical result, Theorem~\ref{theorem:optimization-linear-kernel}, on the Mean Squared Error (MSE) loss of the unlabeled test nodes $\cV_{\cU}$ measured by the gap between  $\bth{\bF(\bW,t)}_{\cU}$ and $\bth{\tilde \bY}_{\cU}$ when using the low-rank feature $\bH$ with $r_0 \in [N]$, which is the generalization error bound for the linear transductive classifier using $\bF(\bW) = \bH \bW$ to predict the labels of the unlabeled nodes. Similar to existing works such as~\citep{KothapalliTB23-neural-collapse} that use the Mean Squared Error (MSE) to analyze the optimization and the generalization of GNNs, we employ the MSE loss to provide the generalization error of the node classifier in the following theorem. It is remarked that the MSE loss is necessary for the generalization analysis of transductive learning using the transductive local Rademacher complexity
\citep{TolstikhinBK14-local-complexity-TRC,TLRC}.
\begin{theorem}
\label{theorem:optimization-linear-kernel}
Let $m \ge cN$ for a constant $c\in (0,1)$, and $r_0 \in [N]$. Assume that a set $\cL$ with $\abth{\cL} = m$ is sampled uniformly without replacement from $[N]$, and the remaining nodes $\cV_U = \cV \setminus \cV_L$ are the test nodes. Then for every $x > 0$, with probability at least $1-\exp(-x)$, after the $t$-th iteration of gradient descent for all $t \ge 1$, we have
\bal\label{eq:optimization-linear-kernel-test-loss}
\resizebox{1\textwidth}{!}
{$
\cU_{\textup{test}}(t) \defeq \frac 1u \fnorm{\bth{\bF(\bW,t) - \tilde \bY}_{\cU}}^2 \le \frac {2c_0}m \pth{L_1(\bK,\tilde \bY,t) + L_2(\bK,\bN,t)}  + c_0\textup{KC}(\bK) + \frac{c_0x}{u},$}
\eal
where $c_0$ is a positive number depending on $\bU$, $\set{\hat \lambda_i}_{i=1}^{r_0}$, and $\tau_0$ with $\tau_0^2 = \max_{i \in [N]} \bK_{ii}$.
$L_1(\bK,\tilde \bY,t) \defeq \fnorm{\pth{\bI_ m- \eta \bth{\bK}_{\cL,\cL} }^t\bth{\tilde \bY}_{\cL}}^2$,
$L_2(\bK,\bN,t) = \fnorm{\eta \bth{\bK}_{\cL,\cL} \sum_{t'=0}^{t-1} \pth{\bI_ m- \eta \bth{\bK}_{\cL,\cL} }^{t'}\bth{\bN}_{\cL}}^2$.
$\textup{KC}$ is the kernel complexity of the gram matrix $\bK = \bH \bH^{\top}$ defined by $\textup{KC}(\bK) = \min_{r_0 \in [N]} r_0\pth{\frac{1}{u} + \frac{1}{m}} +  \sqrt{\norm{\bK}{r_0}} \pth{ \frac{1}{\sqrt u} + \frac{1}{\sqrt m}}$.
\end{theorem}
This theorem is proved in Section~\ref{sec:transductive-theory}. Specifically, $\cU_{\textup{test}}(t)$ denotes the test loss over unlabeled nodes, quantified by the discrepancy between the classifier output $\bF(\bW,t)$ and the clean label matrix $\tilde \bY$. The upper bound on the test loss in (\ref{eq:optimization-linear-kernel-test-loss}) consists of three components: $L_1(\bK,\tilde \bY,t)$, $L_2(\bK,\bN,t)$, and $\text{KC}({\mathbf{K}})$, each serving a distinct role. The term $L_1(\bK,\tilde \bY,t)$ reflects the training loss of the classifier when using clean labels. $L_2(\bK,\bN,t)$ captures the impact of label noise on the classification loss. The term $\text{KC}({\mathbf{K}})$ denotes the kernel complexity (KC), which assesses the structural complexity of the gram matrix $\mathbf{K}$ derived from node representations $\mathbf{H}$ produced by the GCL-LRR encoder. Importantly, the TNN $\norm{\bK}{r_0}$ appears on the right-hand side of the upper bound in (\ref{eq:optimization-linear-kernel-test-loss}), thereby providing theoretical justification for incorporating the TNN regularizer $\norm{\bK}{r_0}$ into the GCL-LRR objective function (\ref{eq:loss-GCL-LR-overall}) to promote low-rank feature learning. Furthermore, under the LFP, which is consistently supported by the empirical evidence shown in Figure~\ref{fig:eigen-projection} and Figure~\ref{fig:eigen-projection-more}, $L_1(\bK,\tilde \bY,t)$ diminishes as the number of training iterations $t$ increases. Simultaneously, $L_2(\bK,\bN,t)$ remains small due to the approximately uniform eigen-projection of label noise, while $\bK = \bH^{\top}\bH$ remains close to a rank-$r_0$ matrix, as the TNN is effectively minimized through the GCL-LRR training objective (\ref{eq:loss-GCL-LR-overall}). In our experiments, described in the subsequent section, we select the TNN rank $r_0$ via standard cross-validation across all graph datasets. As reported in Table~\ref{table:xi-gamma0}, the optimal rank $r_0$ consistently falls within the range of $0.1 \min\set{N,d}$ to $0.3 \min\set{N,d}$.


\subsection{GCL-LR-Attention: Improving GCL-LRR by Low Rank Attention}
\label{sec:GCL-LR-Attention}
To further enhance the performance of GCL-LRR, we introduce an improved model termed GCL-LR-Attention, which incorporates a novel LR-Attention layer into GCL-LRR. GCL-LR-Attention applies self-attention to the low-rank node representations $\bH \in \RR^{N \times d}$ produced by the GCL-LRR encoder via the optimization of (\ref{eq:loss-GCL-LR-overall}). Specifically, the output of the LR-Attention layer of the GCL-LR-Attention model is computed as $\bF = \bB \bH$, where $\bF$ denotes the attention-transformed features, and $\bB \in \RR^{N \times N}$ is the attention weight matrix. We recall that the gram matrix of the node features is given by $\bK = \bH \bH^{\top}$. In our LR-Attention formulation, the attention matrix is defined as $\bB = \bK / {\hat \lambda_1}$, where $\hat \lambda_1$ is the largest eigenvalue of $\bK$.

The resulting gram matrix of the transformed features $\bF \in \RR^{N \times d}$ is $\bK_{\bF} = \bF \bF^{\top} = \bK^3 / {\hat \lambda_1^2}$. Let $\set{\lambda_i}{i=1}^N$ denote the eigenvalues of $\bK{\bF}$, ordered as $\lambda_1 \ge \lambda_2 \ge \ldots \ge \lambda_N \ge 0$. Then, for each $i \in [N]$, it holds that $\lambda_i = \hat \lambda_i^3 / {\hat \lambda_1^2}$. Noting that $\lambda_i = \hat \lambda_i \cdot \hat \lambda_i^2 / {\hat \lambda_1^2} \le \hat \lambda_i$ due to the ordering $\lambda_1 \ge \lambda_i$ for all $i \in [N]$, it follows that the LR-Attention layer reduces the KC of the original gram matrix $\bK$. Therefore, the KC of $\bK_{\bF}$ is guaranteed to be no greater than that of $\bK$, demonstrating the regularization effect introduced by the LR-Attention mechanism. We then train a transductive classifier on top of $\bF$ similar to Section~\ref{sec:low-rank-transductive} by minimizing the
following loss function,
\bal\label{eq:node-classification-obj-LRA}
\min\limits_{\bW} L(\bW) =  \frac{1}{m} \sum_{v_i \in \mathcal{V_L}} \KL \pth{\by_i, \bth{\softmax\pth{\bF \bW}}_i},
\eal
where $\bW$ is the weight matrix for the classifier. The linear classifier trained with the LR-Attention layer via the optimization objective in (\ref{eq:node-classification-obj-LRA}) is referred to as GCL-LR-Attention. Based on the preceding analysis and the test loss upper bound given in Equation (\ref{eq:optimization-linear-kernel-test-loss}) in Theorem~\ref{theorem:optimization-linear-kernel}, GCL-LR-Attention exhibits a reduced KC compared to GCL-LRR. Consequently, the test loss $\cU_{\textup{test}}(t)$ for GCL-LR-Attention can be lower than that of GCL-LRR, implying improved predictive performance. This theoretical advantage is empirically supported by the results in Table~\ref{tab:kernel_complexity} and Table~\ref{tab:upper_bound}, where GCL-LR-Attention consistently achieves both a lower kernel complexity and a tighter upper bound on the test loss compared to GCL-LRR.

\section{Experiments}
\label{sec:exp}
In this section, we comprehensively evaluate the performance of GCL-LRR and GCL-LR-Attention on several public graph datasets. Section~\ref{sec:setting} outlines the experimental setup and implementation details for both models. In Section~\ref{sec:classification}, we report the results of GCL-LRR and GCL-LR-Attention on semi-supervised node classification under various types of label noise. Section~\ref{sec:kernel_complexity} investigates the KC and the theoretical upper bound of the test loss for both GCL-LRR and GCL-LR-Attention. Section~\ref{sec:time-rank} presents a runtime comparison between GCL-LRR, GCL-LR-Attention, and baseline methods. In Section~\ref{sec:rank_study}, we conduct an ablation study to examine the effect of the rank parameter $r_0$ in the TNN. Further analysis of KC across additional datasets is given in Section~\ref{sec:sup_kernel_complexity}, and extended node classification results on more benchmarks are reported in Section~\ref{sec:sup_classification-more-data}. Complementary results on eigen-projection and signal concentration ratios are presented in Section~\ref{sec:sup_projection}.
The effectiveness of GCL-LRR and GCL-LR-Attention on heterophilic graph datasets is studied in Section~\ref{sec:heterophlic}. Section~\ref{sec:trans_classifier} compares our models with existing graph contrastive learning methods that employ various classifier architectures. The statistical significance of improvements observed in Sections~\ref{sec:classification}, \ref{sec:sup_classification-more-data}, and \ref{sec:heterophlic} is assessed using the Student's $t$-test, detailed in Section~\ref{sec:t-test}. Finally, Section~\ref{sec:sensitivity} presents a sensitivity analysis of the key hyperparameter.

\subsection{Experimental Settings}
\label{sec:setting}
In our experiments, we evaluate the proposed methods using eight widely adopted graph benchmark datasets: Cora, Citeseer, and PubMed~\citep{sen_2008_aimag}, Coauthor CS, ogbn-arxiv~\citep{hu2020open}, Wiki-CS~\citep{mernyei2020wiki}, and Amazon-Computers and Amazon-Photos~\citep{shchur2018pitfalls}. Since these public datasets do not include label corruption or attribute noise by default, we introduce noise manually to assess the robustness of our algorithms.
To simulate label corruption, we follow established protocols from prior work~\citep{han2020survey, dai2022towards, qian2022robust} and generate two types of noisy labels: (1) Symmetric noise, where each label is randomly flipped to any other class with equal probability; and (2) Asymmetric noise, where label flips occur preferentially between semantically similar classes. Specifically, we adopt the formal label noise model introduced in~\citep{song2022learning}. Let $\mathbf{T} \in [0,1]^{C \times C}$ denote the noise transition matrix, where $\mathbf{T}_{ij} := \mathbb{P}(\tilde{y} = j \mid y = i)$ defines the probability that a clean label $y = i$ is flipped to a noisy label $\tilde{y} = j$. Under symmetric noise with rate $\tau \in [0,1]$, the transition matrix is defined by $\mathbf{T}_{ii} = 1 - \tau$ and $\mathbf{T}_{ij} = \frac{\tau}{C - 1}$ for all $j \ne i$. For asymmetric noise, the matrix satisfies $\mathbf{T}_{ii} = 1 - \tau$ with $\mathbf{T}_{ij} > \mathbf{T}_{ik}$ for some $j \ne i$, $k \ne i$, capturing the structured mislabeling observed in real-world scenarios.
To assess performance under attribute noise, we follow the approach of~\citep{ding2022data}, wherein a fixed proportion of each node’s input features is randomly shuffled. The ratio of perturbed features defines the attribute noise level used in our experiments.

For all experiments, we adopt the standard dataset splits for training, validation, and testing as established in prior works~\citep{shchur2018pitfalls, mernyei2020wiki, hu2020open}. Noise is introduced exclusively into the training and validation sets, while the test set remains unaltered to ensure unbiased evaluation.
Hyperparameter tuning is conducted via five-fold cross-validation on the training data of each benchmark. Specifically, we explore the learning rate over the range ${1 \times 10^{-4}, 5 \times 10^{-4}, 1 \times 10^{-3}, 5 \times 10^{-3}, 1 \times 10^{-2}, 3 \times 10^{-2}, 6 \times 10^{-2}, 1 \times 10^{-1}, 5 \times 10^{-1}}$ and weight decay from ${1 \times 10^{-5}, 5 \times 10^{-5}, 1 \times 10^{-4}, 5 \times 10^{-4}, 1 \times 10^{-3}, 5 \times 10^{-3}}$. Dropout rates are chosen from ${0.3, 0.4, 0.5, 0.6, 0.7}$. The optimal hyperparameters for each dataset are those that yield the lowest validation loss. All models are trained using the Adam optimizer for up to $500$ epochs, with early stopping triggered if the validation loss fails to improve over $20$ successive epochs. To ensure robustness against initialization variance, each experiment is repeated $10$ times with different random seeds.

\noindent \textbf{Cross-Validation for Tuning $r_0$ and $\tau$.}
We determine the rank parameter $r_0$ and the weight $\tau$ of the TNN loss via cross-validation on each dataset. The rank is set as $r_0 = \lceil \gamma \min\set{N,d} \rceil$, where $\gamma$ denotes the rank ratio. We select $\gamma$ from $\{0.1, 0.2, 0.3, 0.4, 0.5, 0.6, 0.7, 0.8, 0.9\}$ and $\tau$ from $\{0.05, 0.1, 0.15, 0.2, 0.25, 0.3, 0.35, 0.4, 0.45, 0.5\}$ using five-fold cross-validation on $20\%$ of the training set. The chosen values for each dataset are listed in Table~\ref{tab:hyper-parameters}.

\begin{table*}[!htpb]
\small
\centering
\caption{Selected rank ratio $\gamma$ and TNN weight $\lambda$ for each dataset.}
\label{table:xi-gamma0}
\resizebox{1\textwidth}{!}{
\begin{tabular}{|c|cccccccc|}
\hline
Parameters    & Cora & Citeseer & PubMed & Coauthor CS & ogbn-arxiv & Wiki-CS & Amazon-Computers & Amazon-Photos \\ \hline
$\tau$      & 0.10 & 0.10     & 0.10   & 0.20        & 0.10 & 0.25 & 0.20  & 0.20       \\
$\gamma$ & 0.2  & 0.2      & 0.3    & 0.3         & 0.4  & 0.2& 0.2& 0.3      \\ \hline
\end{tabular}
        }
\label{tab:hyper-parameters}
\end{table*}


\noindent \textbf{Datasets.} We evaluate our method on eight public benchmarks that are widely used for node representation learning, namely Cora, Citeseer, PubMed \citep{sen_2008_aimag}, Coauthor CS, ogbn-arxiv \citep{hu2020open}, Wiki-CS~\citep{mernyei2020wiki}, Amazon-Computers, and Amazon-Photos~\citep{shchur2018pitfalls}. Cora, Citeseer, and PubMed are the three most widely used citation networks.
Coauthor CS is a co-authorship graph. The ogbn-arxiv is a directed citation graph.
Wiki-CS is a hyperlink network of computer science articles. Amazon-Computers and Amazon-Photos are co-purchase networks of products selling on Amazon.com.
We summarize the statistics of all the datasets in Table~\ref{tab:dataset}.
\begin{table}[ht]
\footnotesize
\centering
\caption{Statistics of the datasets.}
\label{tab:dataset}
\begin{tabular}{|c|cccc|}
\hline
Dataset & Nodes & Edges & Features & Classes \\
\hline
Cora & 2,708 & 5,429 & 1,433 & 7 \\
CiteSeer & 3,327 & 4,732 & 3,703 & 6 \\
PubMed & 19,717 & 44,338 & 500 & 3 \\
Coauthor CS & 18,333 & 81,894 & 6,805 & 15 \\
ogbn-arxiv & 169,343 & 1,166,243 & 128 & 40 \\
Wiki-CS & 11,701 & 215,863 & 300 & 10 \\
Amazon-Computers & 13,752 & 245,861 & 767 & 10 \\
Amazon-Photos & 7,650 & 119,081 & 745 & 8 \\
\hline
\end{tabular}
\end{table}


\subsection{Node Classification}
\label{sec:classification}

\textbf{Compared Methods.}
We conduct a comprehensive comparison of GCL-LRR and GCL-LR-Attention with various semi-supervised node representation learning approaches, including GCN~\citep{kipf2017semi}, GCE~\citep{zhang2018generalized}, S$^2$GC~\citep{zhu2020simple}, and GRAND+\citep{feng2022grand+}. Additionally, we incorporate two baseline methods specifically designed to handle node classification under label noise: NRGNN\citep{dai2021nrgnn} and RTGNN~\citep{qian2022robust}. To benchmark against leading graph contrastive learning (GCL) techniques, we further evaluate GCL-LRR alongside state-of-the-art methods such as GraphCL~\citep{you2020graph}, MERIT~\citep{Jin2021MultiScaleCS}, SUGRL~\citep{mo2022simple}, and SFA~\citep{zhang2023spectral}.
Moreover, we include attention-based graph neural networks (GNNs) in our comparison, specifically GFSA~\citep{GFSA} and HONGAT~\citep{zhang2024hongat}, both of which are designed to integrate low- and high-frequency information extracted from graph structures. We also compare GCL-LRR with CRGNN~\citep{CRGNN} and CGNN~\citep{CGNN}, which utilize contrastive learning strategies to address noisy labels in graph data.
To further validate the robustness of GCL-LRR in learning effective node representations, we compare it with two contrastive learning frameworks originally proposed for visual data, Jo-SRC~\citep{yao2021jo} and Sel-CL~\citep{li2022selective}, which rely on the selection of clean samples. Given the generality of their selection mechanisms, we adapt both methods to the graph learning setting for our experiments.
Jo-SRC employs the Jensen-Shannon divergence to identify clean training instances via a general representation space selection framework. It further enhances robustness by incorporating a consistency regularization term into the contrastive loss. In our implementation, we adapt Jo-SRC’s sample selection and consistency regularization mechanisms to the state-of-the-art GCL method, MERIT. Specifically, we modify the graph contrastive loss in MERIT to include the regularization term proposed in Jo-SRC and restrict training to only the clean samples identified through Jo-SRC’s selection process. Sel-CL aims to learn robust pre-trained representations by constructing contrastive pairs selectively from confidently labeled examples. These examples are identified based on the alignment between learned representations and propagated labels, evaluated using cross-entropy loss. Sel-CL then filters pairs whose representation similarity exceeds a dynamically computed threshold. In our adaptation, we incorporate Sel-CL’s confident contrastive pair selection strategy into MERIT by selecting high-confidence node pairs in the representation space for contrastive learning.

\begin{table*}[!htb]
\caption{Performance comparison for node classification on Cora, Citeseer, PubMed, and Coauthor-CS with asymmetric label noise, symmetric label noise, and attribute noise. The highest values for each dataset under each setting in the table are bold, and the second-lowest values are underlined. The results represent the mean values computed over $10$ independent runs, with the standard deviation reported after $\pm$.}
\label{table:label_noise_short}
\resizebox{\textwidth}{!}{
\begin{tabular}{|c|c|cccccccccc|}
\hline
  &  &  \multicolumn{10}{c|}{Noise Type}    \\  \cmidrule{3-12}
  &  &  \multicolumn{1}{c|}{0}  &  \multicolumn{3}{c|}{40}  &  \multicolumn{3}{c|}{60}  &  \multicolumn{3}{c|}{80}  \\  \cmidrule{3-12}
\multirow{-3}{*}{Dataset}  &  \multirow{-3}{*}{Methods}  &  \multicolumn{1}{c|}{-}  &  Asymmetric  &  Symmetric  &  \multicolumn{1}{c|}{Attribute}  &  Asymmetric  &  Symmetric  &  \multicolumn{1}{c|}{Attribute}  &  Asymmetric  &  Symmetric  &  Attribute  \\  \hline
\multirow{19}{*}{Cora}  &  GCN   &  \multicolumn{1}{c|}{0.815$\pm$0.005}  &  0.547$\pm$0.015  &  0.636$\pm$0.007  &  \multicolumn{1}{c|}{0.639$\pm$0.008}  &  0.405$\pm$0.014  &  0.517$\pm$0.010  &  \multicolumn{1}{c|}{0.439$\pm$0.012}  &  0.265$\pm$0.012  &  0.354$\pm$0.014  &  0.317$\pm$0.013  \\
  &  S$^2$GC  &  \multicolumn{1}{c|}{0.835$\pm$0.002}  &  0.569$\pm$0.007  &  0.664$\pm$0.007  &  \multicolumn{1}{c|}{0.661$\pm$0.007}  &  0.422$\pm$0.010  &  0.535$\pm$0.010  &  \multicolumn{1}{c|}{0.454$\pm$0.011}  &  0.279$\pm$0.014  &  0.366$\pm$0.014  &  0.320$\pm$0.013  \\
  &  GCE  &  \multicolumn{1}{c|}{0.819$\pm$0.004}  &  0.573$\pm$0.011  &  0.652$\pm$0.008  &  \multicolumn{1}{c|}{0.650$\pm$0.014}  &  0.449$\pm$0.011  &  0.509$\pm$0.011  &  \multicolumn{1}{c|}{0.445$\pm$0.015}  &  0.280$\pm$0.013  &  0.353$\pm$0.013  &  0.325$\pm$0.015  \\
  &  UnionNET  &  \multicolumn{1}{c|}{0.820$\pm$0.006}  &  0.569$\pm$0.014  &  0.664$\pm$0.007  &  \multicolumn{1}{c|}{0.653$\pm$0.012}  &  0.452$\pm$0.010  &  0.541$\pm$0.010  &  \multicolumn{1}{c|}{0.450$\pm$0.009}  &  0.283$\pm$0.014  &  0.370$\pm$0.011  &  0.320$\pm$0.012  \\
  &  NRGNN  &  \multicolumn{1}{c|}{0.822$\pm$0.006}  &  0.571$\pm$0.019  &  0.676$\pm$0.007  &  \multicolumn{1}{c|}{0.645$\pm$0.012}  &  0.470$\pm$0.014  &  0.548$\pm$0.014  &  \multicolumn{1}{c|}{0.451$\pm$0.011}  &  0.282$\pm$0.022  &  0.373$\pm$0.012  &  0.326$\pm$0.010  \\
  &  RTGNN  &  \multicolumn{1}{c|}{0.828$\pm$0.003}  &  0.570$\pm$0.010  &  0.682$\pm$0.008  &  \multicolumn{1}{c|}{0.678$\pm$0.011}  &  0.474$\pm$0.011  &  0.555$\pm$0.010  &  \multicolumn{1}{c|}{0.457$\pm$0.009}  &  0.280$\pm$0.011  &  0.386$\pm$0.014  &  0.342$\pm$0.016  \\
  &  SUGRL  &  \multicolumn{1}{c|}{0.834$\pm$0.005}  &  0.564$\pm$0.011  &  0.674$\pm$0.012  &  \multicolumn{1}{c|}{0.675$\pm$0.009}  &  0.468$\pm$0.011  &  0.552$\pm$0.011  &  \multicolumn{1}{c|}{0.452$\pm$0.012}  &  0.280$\pm$0.012  &  0.381$\pm$0.012  &  0.338$\pm$0.014  \\
  &  MERIT  &  \multicolumn{1}{c|}{0.831$\pm$0.005}  &  0.560$\pm$0.008  &  0.670$\pm$0.008  &  \multicolumn{1}{c|}{0.671$\pm$0.009}  &  0.467$\pm$0.013  &  0.547$\pm$0.013  &  \multicolumn{1}{c|}{0.450$\pm$0.014}  &  0.277$\pm$0.013  &  0.385$\pm$0.013  &  0.335$\pm$0.009  \\
  &  ARIEL  &  \multicolumn{1}{c|}{0.843$\pm$0.004}  &  0.573$\pm$0.013  &  0.681$\pm$0.010  &  \multicolumn{1}{c|}{0.675$\pm$0.009}  &  0.471$\pm$0.012  &  0.553$\pm$0.012  &  \multicolumn{1}{c|}{0.455$\pm$0.014}  &  0.284$\pm$0.014  &  0.389$\pm$0.013  &  0.343$\pm$0.013  \\
  &  SFA  &  \multicolumn{1}{c|}{0.839$\pm$0.010}  &  0.564$\pm$0.011  &  0.677$\pm$0.013  &  \multicolumn{1}{c|}{0.676$\pm$0.015}  &  0.473$\pm$0.014  &  0.549$\pm$0.014  &  \multicolumn{1}{c|}{0.457$\pm$0.014}  &  0.282$\pm$0.016  &  0.389$\pm$0.013  &  0.344$\pm$0.017  \\
  &  Sel-Cl  &  \multicolumn{1}{c|}{0.828$\pm$0.002}  &  0.570$\pm$0.010  &  0.685$\pm$0.012  &  \multicolumn{1}{c|}{0.676$\pm$0.009}  &  0.472$\pm$0.013  &  0.554$\pm$0.014  &  \multicolumn{1}{c|}{0.455$\pm$0.011}  &  0.282$\pm$0.017  &  0.389$\pm$0.013  &  0.341$\pm$0.015  \\
  &  Jo-SRC  &  \multicolumn{1}{c|}{0.825$\pm$0.005}  &  0.571$\pm$0.006  &  0.684$\pm$0.013  &  \multicolumn{1}{c|}{0.679$\pm$0.007}  &  0.473$\pm$0.011  &  0.556$\pm$0.008  &  \multicolumn{1}{c|}{0.458$\pm$0.012}  &  0.285$\pm$0.013  &  0.387$\pm$0.018  &  0.345$\pm$0.018  \\
  &  GRAND+  &  \multicolumn{1}{c|}{0.858$\pm$0.006}  &  0.570$\pm$0.009  &  0.682$\pm$0.007  &  \multicolumn{1}{c|}{0.678$\pm$0.011}  &  0.472$\pm$0.010  &  0.554$\pm$0.008  &  \multicolumn{1}{c|}{0.456$\pm$0.012}  &  0.284$\pm$0.015  &  0.387$\pm$0.015  &  0.345$\pm$0.013  \\
  &  GFSA  &  \multicolumn{1}{c|}{0.837$\pm$0.006}  &  0.568$\pm$0.012  &  0.676$\pm$0.010  &  \multicolumn{1}{c|}{0.672$\pm$0.009}  &  0.466$\pm$0.012  &  0.545$\pm$0.013  &  \multicolumn{1}{c|}{0.451$\pm$0.012}  &  0.279$\pm$0.012  &  0.384$\pm$0.015  &  0.336$\pm$0.013  \\
  &  HONGAT  &  \multicolumn{1}{c|}{0.833$\pm$0.004}  &  0.566$\pm$0.011  &  0.673$\pm$0.011  &  \multicolumn{1}{c|}{0.667$\pm$0.010}  &  0.464$\pm$0.010  &  0.543$\pm$0.011  &  \multicolumn{1}{c|}{0.449$\pm$0.010}  &  0.278$\pm$0.013  &  0.381$\pm$0.014  &  0.334$\pm$0.014  \\
  &  CRGNN  &  \multicolumn{1}{c|}{0.842$\pm$0.005}  &  0.572$\pm$0.010  &  0.678$\pm$0.010  &  \multicolumn{1}{c|}{0.674$\pm$0.010}  &  0.470$\pm$0.012  &  0.551$\pm$0.013  &  \multicolumn{1}{c|}{0.454$\pm$0.013}  &  0.283$\pm$0.014  &  0.386$\pm$0.014  &  0.341$\pm$0.015  \\
  &  CGNN  &  \multicolumn{1}{c|}{0.835$\pm$0.006}  &  0.567$\pm$0.009  &  0.670$\pm$0.012  &  \multicolumn{1}{c|}{0.669$\pm$0.011}  &  0.462$\pm$0.013  &  0.544$\pm$0.011  &  \multicolumn{1}{c|}{0.450$\pm$0.013}  &  0.281$\pm$0.012  &  0.380$\pm$0.013  &  0.337$\pm$0.014  \\
  &  GCL-LRR  &  \multicolumn{1}{c|}{\underline{0.858$\pm$0.006}}  &  \underline{0.589$\pm$0.011}  &  \underline{0.713$\pm$0.007}  &  \multicolumn{1}{c|}{\underline{0.695$\pm$0.011}}  & \underline{0.492$\pm$0.011}  &  \underline{0.587$\pm$0.013}  &  \multicolumn{1}{c|}{\underline{0.477$\pm$0.012}}  &  \underline{0.306$\pm$0.012}  &  \underline{0.419$\pm$0.012}  & \underline{0.363$\pm$0.011}  \\
  & GCL-LR-Attention & \multicolumn{1}{c|}{\textbf{0.861}$\pm$\textbf{0.006}} & \textbf{0.602}$\pm$\textbf{0.011} & \textbf{0.724}$\pm$\textbf{0.007} & \multicolumn{1}{c|}{\textbf{0.708}$\pm$\textbf{0.011}} & \textbf{0.510}$\pm$\textbf{0.011} & \textbf{0.605}$\pm$\textbf{0.013} & \multicolumn{1}{c|}{\textbf{0.492}$\pm$\textbf{0.012}} & \textbf{0.329}$\pm$\textbf{0.012} & \textbf{0.436}$\pm$\textbf{0.012} & \textbf{0.382}$\pm$\textbf{0.011} \\ \hline
\multirow{19}{*}{Citeseer}  &  GCN  &  \multicolumn{1}{c|}{0.703$\pm$0.005}  &  0.475$\pm$0.023  &  0.501$\pm$0.013  &  \multicolumn{1}{c|}{0.529$\pm$0.009}  &  0.351$\pm$0.014  &  0.341$\pm$0.014  &  \multicolumn{1}{c|}{0.372$\pm$0.011}  &  0.291$\pm$0.022  &  0.281$\pm$0.019  &  0.290$\pm$0.014  \\
  &  S$^2$GC  &  \multicolumn{1}{c|}{0.736$\pm$0.005}  &  0.488$\pm$0.013  &  0.528$\pm$0.013  &  \multicolumn{1}{c|}{0.553$\pm$0.008}  &  0.363$\pm$0.012  &  0.367$\pm$0.014  &  \multicolumn{1}{c|}{0.390$\pm$0.013}  &  0.304$\pm$0.024  &  0.284$\pm$0.019  &  0.288$\pm$0.011  \\
  &  GCE  &  \multicolumn{1}{c|}{0.705$\pm$0.004}  &  0.490$\pm$0.016  &  0.512$\pm$0.014  &  \multicolumn{1}{c|}{0.540$\pm$0.014}  &  0.362$\pm$0.015  &  0.352$\pm$0.010  &  \multicolumn{1}{c|}{0.381$\pm$0.009}  &  0.309$\pm$0.012  &  0.285$\pm$0.014  &  0.285$\pm$0.011  \\
  &  UnionNET  &  \multicolumn{1}{c|}{0.706$\pm$0.006}  &  0.499$\pm$0.015  &  0.547$\pm$0.014  &  \multicolumn{1}{c|}{0.545$\pm$0.013}  &  0.379$\pm$0.013  &  0.399$\pm$0.013  &  \multicolumn{1}{c|}{0.379$\pm$0.012}  &  0.322$\pm$0.021  &  0.302$\pm$0.013  &  0.290$\pm$0.012  \\
  &  NRGNN  &  \multicolumn{1}{c|}{0.710$\pm$0.006}  &  0.498$\pm$0.015  &  0.546$\pm$0.015  &  \multicolumn{1}{c|}{0.538$\pm$0.011}  &  0.382$\pm$0.016  &  0.412$\pm$0.016  &  \multicolumn{1}{c|}{0.377$\pm$0.012}  &  0.336$\pm$0.021  &  0.309$\pm$0.018  &  0.284$\pm$0.009  \\
  &  RTGNN  &  \multicolumn{1}{c|}{0.746$\pm$0.008}  &  0.498$\pm$0.007  &  0.556$\pm$0.007  &  \multicolumn{1}{c|}{0.550$\pm$0.012}  &  0.392$\pm$0.010  &  0.424$\pm$0.013  &  \multicolumn{1}{c|}{0.390$\pm$0.014}  &  0.348$\pm$0.017  &  0.308$\pm$0.016  &  0.302$\pm$0.011  \\
  &  SUGRL  &  \multicolumn{1}{c|}{0.730$\pm$0.005}  &  0.493$\pm$0.011  &  0.541$\pm$0.011  &  \multicolumn{1}{c|}{0.544$\pm$0.010}  &  0.376$\pm$0.009  &  0.421$\pm$0.009  &  \multicolumn{1}{c|}{0.388$\pm$0.009}  &  0.339$\pm$0.010  &  0.305$\pm$0.010  &  0.300$\pm$0.009  \\
  &  MERIT  &  \multicolumn{1}{c|}{0.740$\pm$0.007}  &  0.496$\pm$0.012  &  0.536$\pm$0.012  &  \multicolumn{1}{c|}{0.542$\pm$0.010}  &  0.383$\pm$0.011  &  0.425$\pm$0.011  &  \multicolumn{1}{c|}{0.387$\pm$0.008}  &  0.344$\pm$0.014  &  0.301$\pm$0.014  &  0.295$\pm$0.009  \\
  &  SFA  &  \multicolumn{1}{c|}{0.740$\pm$0.011}  &  0.502$\pm$0.014  &  0.532$\pm$0.015  &  \multicolumn{1}{c|}{0.547$\pm$0.013}  &  0.390$\pm$0.014  &  0.433$\pm$0.014  &  \multicolumn{1}{c|}{0.389$\pm$0.012}  &  0.347$\pm$0.016  &  0.312$\pm$0.015  &  0.299$\pm$0.013  \\
  &  ARIEL  &  \multicolumn{1}{c|}{0.727$\pm$0.007}  &  0.500$\pm$0.008  &  0.550$\pm$0.013  &  \multicolumn{1}{c|}{0.548$\pm$0.008}  &  0.391$\pm$0.009  &  0.427$\pm$0.012  &  \multicolumn{1}{c|}{0.389$\pm$0.014}  &  0.349$\pm$0.014  &  0.307$\pm$0.013  &  0.299$\pm$0.013  \\
  &  Sel-Cl  &  \multicolumn{1}{c|}{0.725$\pm$0.008}  &  0.499$\pm$0.012  &  0.551$\pm$0.010  &  \multicolumn{1}{c|}{0.549$\pm$0.008}  &  0.389$\pm$0.011  &  0.426$\pm$0.008  &  \multicolumn{1}{c|}{0.391$\pm$0.020}  &  0.350$\pm$0.018  &  0.310$\pm$0.015  &  0.300$\pm$0.017  \\
  &  Jo-SRC  &  \multicolumn{1}{c|}{0.730$\pm$0.005}  &  0.500$\pm$0.013  &  0.555$\pm$0.011  &  \multicolumn{1}{c|}{0.551$\pm$0.011}  &  0.394$\pm$0.013  &  0.425$\pm$0.013  &  \multicolumn{1}{c|}{0.393$\pm$0.013}  &  0.351$\pm$0.013  &  0.305$\pm$0.018  &  0.303$\pm$0.013  \\
  &  GRAND+  &  \multicolumn{1}{c|}{0.756$\pm$0.004}  &  0.497$\pm$0.010  &  0.553$\pm$0.010  &  \multicolumn{1}{c|}{0.552$\pm$0.011}  &  0.390$\pm$0.013  &  0.422$\pm$0.013  &  \multicolumn{1}{c|}{0.387$\pm$0.013}  &  0.348$\pm$0.013  &  0.309$\pm$0.014  &  0.302$\pm$0.012  \\
  &  GFSA  &  \multicolumn{1}{c|}{0.743$\pm$0.006}  &  0.495$\pm$0.012  &  0.546$\pm$0.012  &  \multicolumn{1}{c|}{0.546$\pm$0.011}  &  0.386$\pm$0.011  &  0.418$\pm$0.011  &  \multicolumn{1}{c|}{0.386$\pm$0.012}  &  0.342$\pm$0.013  &  0.308$\pm$0.015  &  0.298$\pm$0.012  \\
  &  HONGAT  &  \multicolumn{1}{c|}{0.738$\pm$0.007}  &  0.492$\pm$0.014  &  0.540$\pm$0.011  &  \multicolumn{1}{c|}{0.545$\pm$0.009}  &  0.380$\pm$0.012  &  0.413$\pm$0.010  &  \multicolumn{1}{c|}{0.384$\pm$0.013}  &  0.340$\pm$0.014  &  0.306$\pm$0.016  &  0.296$\pm$0.011  \\
  &  CRGNN  &  \multicolumn{1}{c|}{0.751$\pm$0.006}  &  0.497$\pm$0.011  &  0.552$\pm$0.010  &  \multicolumn{1}{c|}{0.549$\pm$0.012}  &  0.389$\pm$0.014  &  0.423$\pm$0.013  &  \multicolumn{1}{c|}{0.388$\pm$0.012}  &  0.347$\pm$0.015  &  0.310$\pm$0.014  &  0.301$\pm$0.012  \\
  &  CGNN  &  \multicolumn{1}{c|}{0.741$\pm$0.007}  &  0.493$\pm$0.013  &  0.544$\pm$0.012  &  \multicolumn{1}{c|}{0.546$\pm$0.010}  &  0.385$\pm$0.013  &  0.419$\pm$0.012  &  \multicolumn{1}{c|}{0.385$\pm$0.011}  &  0.343$\pm$0.013  &  0.307$\pm$0.013  &  0.297$\pm$0.012  \\
  &  GCL-LRR  &  \multicolumn{1}{c|}{\underline{0.757$\pm$0.010}}  &  \underline{0.520$\pm$0.013}  &  \underline{0.581$\pm$0.013}  &  \multicolumn{1}{c|}{\underline{0.570$\pm$0.007}}  &  \underline{0.410$\pm$0.014}  &  \underline{0.455$\pm$0.014}  &  \multicolumn{1}{c|}{\underline{0.406$\pm$0.012}}  &  \underline{0.369$\pm$0.012}  &  \underline{0.335$\pm$0.014}  &  \underline{0.318$\pm$0.010}  \\
  & GCL-LR-Attention & \multicolumn{1}{c|}{\textbf{0.762}$\pm$\textbf{0.010}} & \textbf{0.533}$\pm$\textbf{0.013} & \textbf{0.597}$\pm$\textbf{0.013} & \multicolumn{1}{c|}{\textbf{0.588}$\pm$\textbf{0.007}} & \textbf{0.430}$\pm$\textbf{0.014} & \textbf{0.472}$\pm$\textbf{0.014} & \multicolumn{1}{c|}{\textbf{0.423}$\pm$\textbf{0.012}} & \textbf{0.392}$\pm$\textbf{0.012} & \textbf{0.352}$\pm$\textbf{0.014} & \textbf{0.335}$\pm$\textbf{0.010} \\ \hline
\multirow{19}{*}{PubMed}  &  GCN  &  \multicolumn{1}{c|}{0.790$\pm$0.007}  &  0.584$\pm$0.022  &  0.574$\pm$0.012  &  \multicolumn{1}{c|}{0.595$\pm$0.012}  &  0.405$\pm$0.025  &  0.386$\pm$0.011  &  \multicolumn{1}{c|}{0.488$\pm$0.013}  &  0.305$\pm$0.022  &  0.295$\pm$0.013  &  0.423$\pm$0.013  \\
  &  S$^2$GC  &  \multicolumn{1}{c|}{0.802$\pm$0.005}  &  0.585$\pm$0.023  &  0.589$\pm$0.013  &  \multicolumn{1}{c|}{0.610$\pm$0.009}  &  0.421$\pm$0.030  &  0.401$\pm$0.014  &  \multicolumn{1}{c|}{0.497$\pm$0.012}  &  0.310$\pm$0.039  &  0.290$\pm$0.019  &  0.431$\pm$0.010  \\
  &  GCE  &  \multicolumn{1}{c|}{0.792$\pm$0.009}  &  0.589$\pm$0.018  &  0.581$\pm$0.011  &  \multicolumn{1}{c|}{0.590$\pm$0.014}  &  0.430$\pm$0.012  &  0.399$\pm$0.012  &  \multicolumn{1}{c|}{0.491$\pm$0.010}  &  0.311$\pm$0.021  &  0.301$\pm$0.011  &  0.424$\pm$0.012  \\
  &  UnionNET  &  \multicolumn{1}{c|}{0.793$\pm$0.008}  &  0.603$\pm$0.020  &  0.620$\pm$0.012  &  \multicolumn{1}{c|}{0.592$\pm$0.012}  &  0.445$\pm$0.022  &  0.424$\pm$0.013  &  \multicolumn{1}{c|}{0.489$\pm$0.015}  &  0.313$\pm$0.025  &  0.327$\pm$0.015  &  0.435$\pm$0.009  \\
  &  NRGNN  &  \multicolumn{1}{c|}{0.797$\pm$0.008}  &  0.602$\pm$0.022  &  0.618$\pm$0.013  &  \multicolumn{1}{c|}{0.603$\pm$0.008}  &  0.443$\pm$0.012  &  0.434$\pm$0.012  &  \multicolumn{1}{c|}{0.499$\pm$0.009}  &  0.330$\pm$0.023  &  0.325$\pm$0.013  &  0.433$\pm$0.011  \\
  &  RTGNN  &  \multicolumn{1}{c|}{0.797$\pm$0.004}  &  0.610$\pm$0.008  &  0.622$\pm$0.010  &  \multicolumn{1}{c|}{0.614$\pm$0.012}  &  0.455$\pm$0.010  &  0.455$\pm$0.011  &  \multicolumn{1}{c|}{0.501$\pm$0.011}  &  0.335$\pm$0.013  &  0.338$\pm$0.017  &  0.452$\pm$0.013  \\
  &  SUGRL  &  \multicolumn{1}{c|}{0.819$\pm$0.005}  &  0.603$\pm$0.013  &  0.615$\pm$0.013  &  \multicolumn{1}{c|}{0.615$\pm$0.010}  &  0.445$\pm$0.011  &  0.441$\pm$0.011  &  \multicolumn{1}{c|}{0.501$\pm$0.007}  &  0.321$\pm$0.009  &  0.321$\pm$0.009  &  0.446$\pm$0.010  \\
  &  MERIT  &  \multicolumn{1}{c|}{0.801$\pm$0.004}  &  0.593$\pm$0.011  &  0.612$\pm$0.011  &  \multicolumn{1}{c|}{0.613$\pm$0.011}  &  0.447$\pm$0.012  &  0.443$\pm$0.012  &  \multicolumn{1}{c|}{0.497$\pm$0.009}  &  0.328$\pm$0.011  &  0.323$\pm$0.011  &  0.445$\pm$0.009  \\
  &  ARIEL  &  \multicolumn{1}{c|}{0.800$\pm$0.003}  &  0.610$\pm$0.013  &  0.622$\pm$0.010  &  \multicolumn{1}{c|}{0.615$\pm$0.011}  &  0.453$\pm$0.012  &  0.453$\pm$0.012  &  \multicolumn{1}{c|}{0.502$\pm$0.014}  &  0.331$\pm$0.014  &  0.336$\pm$0.018  &  0.457$\pm$0.013  \\
  &  SFA  &  \multicolumn{1}{c|}{0.804$\pm$0.010}&  0.596$\pm$0.011  &  0.615$\pm$0.011  &  \multicolumn{1}{c|}{0.609$\pm$0.011}  &  0.447$\pm$0.014  &  0.446$\pm$0.017  &  \multicolumn{1}{c|}{0.499$\pm$0.014}  &  0.330$\pm$0.011  &  0.327$\pm$0.011  &  0.447$\pm$0.014  \\
  &  Sel-Cl  &  \multicolumn{1}{c|}{0.799$\pm$0.005}  &  0.605$\pm$0.014  &  0.625$\pm$0.012  &  \multicolumn{1}{c|}{0.614$\pm$0.012}  &  0.455$\pm$0.014  &  0.449$\pm$0.010  &  \multicolumn{1}{c|}{0.502$\pm$0.008}  &  0.334$\pm$0.021  &  0.332$\pm$0.014  &  0.456$\pm$0.014  \\
  &  Jo-SRC  &  \multicolumn{1}{c|}{0.801$\pm$0.005}  &  0.613$\pm$0.010  &  0.624$\pm$0.013  &  \multicolumn{1}{c|}{0.617$\pm$0.013}  &  0.453$\pm$0.008  &  0.455$\pm$0.013  &  \multicolumn{1}{c|}{0.504$\pm$0.013}  &  0.330$\pm$0.015  &  0.334$\pm$0.018  &  0.459$\pm$0.018  \\
  &  GRAND+  &  \multicolumn{1}{c|}{0.845$\pm$0.006}  &  0.610$\pm$0.011  &  0.624$\pm$0.013  &  \multicolumn{1}{c|}{0.617$\pm$0.013}  &  0.453$\pm$0.008  &  0.453$\pm$0.011  &  \multicolumn{1}{c|}{0.503$\pm$0.010}  &  0.331$\pm$0.014  &  0.337$\pm$0.013  &  0.458$\pm$0.014  \\
  &  GFSA  &  \multicolumn{1}{c|}{0.823$\pm$0.005}  &  0.608$\pm$0.012  &  0.621$\pm$0.011  &  \multicolumn{1}{c|}{0.616$\pm$0.009}  &  0.450$\pm$0.013  &  0.452$\pm$0.012  &  \multicolumn{1}{c|}{0.500$\pm$0.010}  &  0.333$\pm$0.013  &  0.334$\pm$0.011  &  0.455$\pm$0.012  \\
  &  HONGAT  &  \multicolumn{1}{c|}{0.818$\pm$0.006}  &  0.606$\pm$0.011  &  0.619$\pm$0.012  &  \multicolumn{1}{c|}{0.613$\pm$0.010}  &  0.448$\pm$0.014  &  0.447$\pm$0.012  &  \multicolumn{1}{c|}{0.498$\pm$0.012}  &  0.328$\pm$0.012  &  0.326$\pm$0.013  &  0.450$\pm$0.011  \\
  &  CRGNN  &  \multicolumn{1}{c|}{0.829$\pm$0.005}  &  0.612$\pm$0.010  &  0.623$\pm$0.009  &  \multicolumn{1}{c|}{0.618$\pm$0.011}  &  0.452$\pm$0.011  &  0.455$\pm$0.013  &  \multicolumn{1}{c|}{0.503$\pm$0.009}  &  0.335$\pm$0.013  &  0.333$\pm$0.014  &  0.457$\pm$0.012  \\
  &  CGNN  &  \multicolumn{1}{c|}{0.822$\pm$0.006}  &  0.607$\pm$0.013  &  0.620$\pm$0.011  &  \multicolumn{1}{c|}{0.615$\pm$0.010}  &  0.449$\pm$0.012  &  0.451$\pm$0.014  &  \multicolumn{1}{c|}{0.499$\pm$0.010}  &  0.332$\pm$0.014  &  0.330$\pm$0.012  &  0.454$\pm$0.013  \\
  &  GCL-LRR  &  \multicolumn{1}{c|}{\underline{0.845$\pm$0.009}}  & \underline{0.637$\pm$0.014}  & \underline{0.645$\pm$0.015}  &  \multicolumn{1}{c|}{\underline{0.637$\pm$0.011}}  & \underline{0.479$\pm$0.011}  &  \underline{0.484$\pm$0.013}  &  \multicolumn{1}{c|}{\underline{0.526$\pm$0.011}}  &  \underline{0.356$\pm$0.011}  &  \underline{0.360$\pm$0.012}  &  \underline{0.482$\pm$0.014}  \\
  & GCL-LR-Attention & \multicolumn{1}{c|}{\textbf{0.846}$\pm$\textbf{0.009}} & \textbf{0.652}$\pm$\textbf{0.014} & \textbf{0.662}$\pm$\textbf{0.015} & \multicolumn{1}{c|}{\textbf{0.655}$\pm$\textbf{0.011}} & \textbf{0.498}$\pm$\textbf{0.011} & \textbf{0.503}$\pm$\textbf{0.013} & \multicolumn{1}{c|}{\textbf{0.544}$\pm$\textbf{0.011}} & \textbf{0.379}$\pm$\textbf{0.011} & \textbf{0.379}$\pm$\textbf{0.012} & \textbf{0.498}$\pm$\textbf{0.014} \\ \hline
\multirow{19}{*}{Coauthor-CS}  &  GCN  &  \multicolumn{1}{c|}{0.918$\pm$0.001}  &  0.645$\pm$0.009  &  0.656$\pm$0.006  &  \multicolumn{1}{c|}{0.702$\pm$0.010}  &  0.511$\pm$0.013  &  0.501$\pm$0.009  &  \multicolumn{1}{c|}{0.531$\pm$0.010}  &  0.429$\pm$0.022  &  0.389$\pm$0.011  &  0.415$\pm$0.013  \\
  &  S$^2$GC  &  \multicolumn{1}{c|}{0.918$\pm$0.001}  &  0.657$\pm$0.012  &  0.663$\pm$0.006  &  \multicolumn{1}{c|}{0.713$\pm$0.010}  &  0.516$\pm$0.013  &  0.514$\pm$0.009  &  \multicolumn{1}{c|}{0.556$\pm$0.009}  &  0.437$\pm$0.020  &  0.396$\pm$0.010  &  0.422$\pm$0.012  \\
  &  GCE  &  \multicolumn{1}{c|}{0.922$\pm$0.003}  &  0.662$\pm$0.017  &  0.659$\pm$0.007  &  \multicolumn{1}{c|}{0.705$\pm$0.014}  &  0.515$\pm$0.016  &  0.502$\pm$0.007  &  \multicolumn{1}{c|}{0.539$\pm$0.009}  &  0.443$\pm$0.017  &  0.389$\pm$0.012  &  0.412$\pm$0.011  \\
  &  UnionNET  &  \multicolumn{1}{c|}{0.918$\pm$0.002}  &  0.669$\pm$0.023  &  0.671$\pm$0.013  &  \multicolumn{1}{c|}{0.706$\pm$0.012}  &  0.525$\pm$0.011  &  0.529$\pm$0.011  &  \multicolumn{1}{c|}{0.540$\pm$0.012}  &  0.458$\pm$0.015  &  0.401$\pm$0.011  &  0.420$\pm$0.007  \\
  &  NRGNN  &  \multicolumn{1}{c|}{0.919$\pm$0.002}  &  0.678$\pm$0.014  &  0.689$\pm$0.009  &  \multicolumn{1}{c|}{0.705$\pm$0.012}  &  0.545$\pm$0.021  &  0.556$\pm$0.011  &  \multicolumn{1}{c|}{0.546$\pm$0.011}  &  0.461$\pm$0.012  &  0.410$\pm$0.012  &  0.417$\pm$0.007  \\
  &  RTGNN  &  \multicolumn{1}{c|}{0.920$\pm$0.005}  &  0.678$\pm$0.012  &  0.691$\pm$0.009  &  \multicolumn{1}{c|}{0.712$\pm$0.008}  &  0.559$\pm$0.010  &  0.569$\pm$0.011  &  \multicolumn{1}{c|}{0.560$\pm$0.008}  &  0.455$\pm$0.015  &  0.415$\pm$0.015  &  0.412$\pm$0.014  \\
  &  SUGRL  &  \multicolumn{1}{c|}{0.922$\pm$0.005}  &  0.675$\pm$0.010  &  0.695$\pm$0.010  &  \multicolumn{1}{c|}{0.714$\pm$0.006}  &  0.550$\pm$0.011  &  0.560$\pm$0.011  &  \multicolumn{1}{c|}{0.561$\pm$0.007}  &  0.449$\pm$0.011  &  0.411$\pm$0.011  &  0.429$\pm$0.008  \\
  &  MERIT  &  \multicolumn{1}{c|}{0.924$\pm$0.004}  &  0.679$\pm$0.011  &  0.689$\pm$0.008  &  \multicolumn{1}{c|}{0.709$\pm$0.005}  &  0.552$\pm$0.014  &  0.562$\pm$0.014  &  \multicolumn{1}{c|}{0.562$\pm$0.011}  &  0.452$\pm$0.013  &  0.403$\pm$0.013  &  0.426$\pm$0.005  \\
  &  ARIEL  &  \multicolumn{1}{c|}{0.925$\pm$0.004}  &  0.682$\pm$0.011  &  0.699$\pm$0.009  &  \multicolumn{1}{c|}{0.712$\pm$0.005}  &  0.555$\pm$0.011  &  0.566$\pm$0.011  &  \multicolumn{1}{c|}{0.556$\pm$0.011}  &  0.454$\pm$0.014  &  0.415$\pm$0.019  &  0.427$\pm$0.013  \\
  &  SFA  &  \multicolumn{1}{c|}{0.925$\pm$0.009}&  0.682$\pm$0.011  &  0.690$\pm$0.012  &  \multicolumn{1}{c|}{0.715$\pm$0.012}  &  0.555$\pm$0.015  &  0.567$\pm$0.014  &  \multicolumn{1}{c|}{0.565$\pm$0.013}  &  0.458$\pm$0.013  &  0.402$\pm$0.013  &  0.429$\pm$0.015  \\
  &  Sel-Cl  &  \multicolumn{1}{c|}{0.922$\pm$0.008}  &  0.684$\pm$0.009  &  0.694$\pm$0.012  &  \multicolumn{1}{c|}{0.714$\pm$0.010}  &  0.557$\pm$0.013  &  0.568$\pm$0.013  &  \multicolumn{1}{c|}{0.566$\pm$0.010}  &  0.457$\pm$0.013  &  0.412$\pm$0.017  &  0.425$\pm$0.009  \\
  &  Jo-SRC  &  \multicolumn{1}{c|}{0.921$\pm$0.005}  &  0.684$\pm$0.011  &  0.695$\pm$0.004  &  \multicolumn{1}{c|}{0.709$\pm$0.007}  &  0.560$\pm$0.011  &  0.566$\pm$0.011  &  \multicolumn{1}{c|}{0.561$\pm$0.009}  &  0.456$\pm$0.013  &  0.410$\pm$0.018  &  0.428$\pm$0.010  \\
  &  GRAND+  &  \multicolumn{1}{c|}{0.927$\pm$0.004}  &  0.682$\pm$0.011  &  0.693$\pm$0.006  &  \multicolumn{1}{c|}{0.715$\pm$0.008}  &  0.554$\pm$0.008  &  0.568$\pm$0.013  &  \multicolumn{1}{c|}{0.557$\pm$0.011}  &  0.455$\pm$0.012  &  0.416$\pm$0.013  &  0.428$\pm$0.011  \\
  &  GFSA  &  \multicolumn{1}{c|}{0.923$\pm$0.004}  &  0.679$\pm$0.010  &  0.687$\pm$0.009  &  \multicolumn{1}{c|}{0.711$\pm$0.009}  &  0.550$\pm$0.012  &  0.559$\pm$0.011  &  \multicolumn{1}{c|}{0.558$\pm$0.010}  &  0.453$\pm$0.014  &  0.410$\pm$0.012  &  0.426$\pm$0.011  \\
  &  HONGAT  &  \multicolumn{1}{c|}{0.924$\pm$0.003}  &  0.681$\pm$0.012  &  0.692$\pm$0.010  &  \multicolumn{1}{c|}{0.713$\pm$0.008}  &  0.553$\pm$0.013  &  0.563$\pm$0.013  &  \multicolumn{1}{c|}{0.560$\pm$0.012}  &  0.456$\pm$0.013  &  0.411$\pm$0.015  &  0.427$\pm$0.010  \\
  &  CRGNN  &  \multicolumn{1}{c|}{0.926$\pm$0.005}  &  0.683$\pm$0.011  &  0.690$\pm$0.011  &  \multicolumn{1}{c|}{0.712$\pm$0.007}  &  0.551$\pm$0.015  &  0.561$\pm$0.012  &  \multicolumn{1}{c|}{0.559$\pm$0.011}  &  0.454$\pm$0.012  &  0.412$\pm$0.014  &  0.426$\pm$0.012  \\
  &  CGNN  &  \multicolumn{1}{c|}{0.925$\pm$0.006}  &  0.680$\pm$0.012  &  0.689$\pm$0.012  &  \multicolumn{1}{c|}{0.710$\pm$0.010}  &  0.549$\pm$0.014  &  0.560$\pm$0.012  &  \multicolumn{1}{c|}{0.557$\pm$0.012}  &  0.452$\pm$0.013  &  0.409$\pm$0.015  &  0.425$\pm$0.012  \\
  &  GCL-LRR  &  \multicolumn{1}{c|}{\underline{0.933$\pm$0.006}}  &  \underline{0.699$\pm$0.015}  &  \underline{0.721$\pm$0.011}  &  \multicolumn{1}{c|}{\underline{0.742$\pm$0.015}}  & \underline{0.575$\pm$0.014}  &  \underline{0.595$\pm$0.018}  &  \multicolumn{1}{c|}{\underline{0.588$\pm$0.015}}  &  \underline{0.469$\pm$0.015}  &  \underline{0.438$\pm$0.015}  &  \underline{0.453$\pm$0.017}  \\
  & GCL-LR-Attention & \multicolumn{1}{c|}{\textbf{0.934}$\pm$\textbf{0.006}} & \textbf{0.714}$\pm$\textbf{0.015} & \textbf{0.736}$\pm$\textbf{0.011} & \multicolumn{1}{c|}{\textbf{0.758}$\pm$\textbf{0.015}} & \textbf{0.594}$\pm$\textbf{0.014} & \textbf{0.612}$\pm$\textbf{0.018} & \multicolumn{1}{c|}{\textbf{0.606}$\pm$\textbf{0.015}} & \textbf{0.489}$\pm$\textbf{0.015} & \textbf{0.453}$\pm$\textbf{0.015} & \textbf{0.470}$\pm$\textbf{0.017} \\ \hline\end{tabular}
}
\end{table*}

\noindent\textbf{Experimental Results.}
To evaluate the robustness of GCL-LRR and its variant GCL-LR-Attention, we perform a series of experiments on graphs subjected to both symmetric and asymmetric label noise, with corruption levels varying from $40\%$ to $80\%$ in increments of $20\%$. In parallel, we investigate the effect of attribute noise under the same range and step size. Unlike prior GCL methods such as MERIT, SUGRL, and SFA, which adopt an inductive classification framework by training a linear classifier on node embeddings obtained via contrastive learning, GCL-LRR is evaluated under a transductive setting that incorporates test data during training.
To ensure fair comparisons, all baseline models are retrained using the transductive classifier employed in GCL-LRR, along with a standard two-layer GCN transductive classifier. Details on the performance under different classifier configurations are provided in Section~\ref{sec:trans_classifier}. For methods initially designed for inductive learning, we report their best outcomes among the default inductive classifier and the two transductive classifiers.
Table~\ref{table:label_noise_short} summarizes the results for Cora, Citeseer, PubMed, and Coauthor-CS, presenting the average accuracy and standard deviation across $10$ runs for GCL-LRR, GCL-LR-Attention, and competing baselines. Additional results for node classification under symmetric label noise, asymmetric label noise, and attribute noise on ogbn-arxiv, Wiki-CS, Amazon-Computers, and Amazon-Photos can be found in Table~\ref{table:label_noise_all_2} in Section~\ref{sec:sup_classification-more-data}. Across all datasets and noise conditions, GCL-LRR consistently achieves superior performance over all baselines.
Moreover, the introduction of low-rank attention in GCL-LR-Attention leads to further gains. For instance, under $80\%$ symmetric label noise on PubMed, GCL-LRR surpasses the strongest baseline, RTGNN, by $2.2\%$ in classification accuracy, while GCL-LR-Attention achieves an additional $1.9\%$ improvement. These results underscore the effectiveness of both low-rank regularization and low-rank attention in enhancing the robustness of node classification models.




\subsection{Study in the Kernel Complexity and the Upper Bound of the Test Loss}
\label{sec:kernel_complexity}
This section provides a comparative analysis of each component in the upper bound of the test loss presented in Equation~(\ref{eq:optimization-linear-kernel-test-loss}), specifically $L_1(\bK,\tilde \bY,t)$, $L_2(\bK,\bN,t)$, and $\text{KC}({\mathbf{K}})$, for node representations produced by different methods. The corresponding results are reported in Table~\ref{tab:upper_bound}. Experiments are conducted on Cora, Citeseer, and PubMed under symmetric label noise at a corruption level of $40\%$.
The findings indicate that GCL-LRR and GCL-LR-Attention consistently yield substantially lower values across all evaluated terms when compared to baseline approaches. This suggests superior generalization capacity of these methods in semi-supervised node classification, even in the presence of noisy labels.
Additionally, we evaluate the KC of the gram matrix derived from node representations generated by GCL-LRR, GCL-LR-Attention, and other competitive GCL methods across additional datasets. The results in Table~\ref{tab:kernel_complexity} in Section~\ref{sec:sup_kernel_complexity} show that the representations from GCL-LRR exhibit notably lower complexity. This implies that transductive classifiers trained on such representations tend to incur smaller generalization errors on unlabeled nodes.

\begin{table*}[!htbp]
\centering
\caption{Comparisons on $L_1(\bK,\tilde \bY,t)$, $L_2(\bK,\bN,t)$, $\textup{KC}(\bK)$ and the value of the upper bound of the test loss from Theorem~\ref{theorem:optimization-linear-kernel}. The evaluation is performed on semi-supervised node classification with $40\%$ of symmetric label noise. The lowest values for each dataset in the table are bold, and the second-lowest values are underlined. The results represent the mean values computed over $10$ independent runs, with the standard deviation reported after $\pm$.}
    \label{tab:upper_bound}
\resizebox{1\textwidth}{!}{
\begin{tabular}{|c|c|cccccccc|}
\hline
Datasets&                    & MERIT & SFA & Jo-SRC &GCN &GFSA &HONGAT  & GCL-LRR & GCL-LR-Attention\\ \hline
\multirow{4}{*}{Cora}     & $L_1$       & 5.24 $\pm$0.49 & 6.04 $\pm$0.23 & 6.50  $\pm$0.34 & 7.38 $\pm$0.12 & 6.44 $\pm$0.01 & 6.38  $\pm$ 0.13 & \underline{3.72} $\pm$ 0.38  & \textbf{3.65} $\pm$ 0.38     \\
                          & $L_2$       & 4.92 $\pm$0.14 & 4.95 $\pm$0.35 & 5.05  $\pm$0.13 & 5.24 $\pm$0.01 & 3.80 $\pm$0.24 & 4.25  $\pm$ 0.26 & \underline{2.97} $\pm$ 0.45  & \textbf{2.72} $\pm$ 0.42     \\
                          & KC          & 0.37 $\pm$0.29 & 0.42 $\pm$0.09 & 0.48  $\pm$0.39 & 0.44 $\pm$0.40 & 0.35 $\pm$0.31 & 0.40  $\pm$ 0.08 & \underline{0.20} $\pm$ 0.02  & \textbf{0.18} $\pm$ 0.26     \\
                          & Upper Bound & 10.68$\pm$0.14 & 11.59$\pm$0.15 & 12.18 $\pm$0.46 & 13.22$\pm$0.11 & 10.80$\pm$0.22 & 11.25 $\pm$ 0.02 & \underline{7.05} $\pm$ 0.43  & \textbf{6.74} $\pm$ 0.32     \\ \hline
\multirow{4}{*}{Citeseer} & $L_1$       & 4.72 $\pm$0.42 & 4.85 $\pm$0.28 & 4.92  $\pm$0.23 & 5.10 $\pm$0.40 & 4.54 $\pm$0.46 & 4.69  $\pm$ 0.19 & \underline{4.02} $\pm$ 0.34  & \textbf{3.95} $\pm$ 0.21     \\
                          & $L_2$       & 4.33 $\pm$0.04 & 4.69 $\pm$0.07 & 4.42  $\pm$0.15 & 5.08 $\pm$0.25 & 4.20 $\pm$0.00 & 4.42  $\pm$ 0.03 & \underline{3.75} $\pm$ 0.17  & \textbf{3.60} $\pm$ 0.22     \\
                          & KC          & 0.47 $\pm$0.27 & 0.45 $\pm$0.18 & 0.55  $\pm$0.08 & 0.64 $\pm$0.42 & 0.47 $\pm$0.10 & 0.50  $\pm$ 0.42 & \underline{0.24} $\pm$ 0.18  & \textbf{0.21} $\pm$ 0.16     \\
                          & Upper Bound & 9.77 $\pm$0.14 & 10.21$\pm$0.28 & 10.17 $\pm$0.34 & 11.07$\pm$0.24 & 9.40 $\pm$0.25 & 9.84  $\pm$ 0.14 & \underline{8.20} $\pm$ 0.04  & \textbf{7.97} $\pm$ 0.33     \\ \hline
\multirow{4}{*}{PubMed}   & $L_1$       & 3.97 $\pm$0.29 & 4.02 $\pm$0.08 & 4.11  $\pm$0.14 & 4.35 $\pm$0.06 & 4.26 $\pm$0.12 & 3.95  $\pm$ 0.23 & \underline{3.38} $\pm$ 0.40  & \textbf{3.40} $\pm$ 0.38     \\
                          & $L_2$       & 2.69 $\pm$0.20 & 2.54 $\pm$0.28 & 2.60  $\pm$0.32 & 2.88 $\pm$0.08 & 2.98 $\pm$0.09 & 2.85  $\pm$ 0.03 & \underline{2.32} $\pm$ 0.10  & \textbf{2.26} $\pm$ 0.45     \\
                          & KC          & 0.54 $\pm$0.49 & 0.50 $\pm$0.27 & 0.62  $\pm$0.17 & 0.71 $\pm$0.23 & 0.52 $\pm$0.17 & 0.66  $\pm$ 0.16 & \underline{0.30} $\pm$ 0.29  & \textbf{0.28} $\pm$ 0.37     \\
                          & Upper Bound & 7.44 $\pm$0.22 & 7.28 $\pm$0.03 & 7.59  $\pm$0.37 & 8.15 $\pm$0.26 & 7.99 $\pm$0.23 & 7.63  $\pm$ 0.14 & \underline{6.25} $\pm$ 0.30  & \textbf{6.16} $\pm$ 0.40     \\ \hline

\end{tabular}
}
\end{table*}

We further investigate the performance of GCL-LRR and GCL-LR-Attention on heterophilic graphs, as detailed in Section~\ref{sec:heterophlic}. The node classification results, presented in Table~\ref{tab:heterophlic}, demonstrate that both GCL-LRR and GCL-LR-Attention maintain strong effectiveness in the presence of label noise and attribute noise, even under the structural challenges posed by heterophily.

\subsection{Training Time Comparison}
\label{sec:time-rank}
In this section, we report a comparative analysis of the training time for GCL-LRR and other baseline methods across all benchmark datasets. The total training time for GCL-LRR encompasses three components: the time required for robust graph contrastive learning, the computation time for the singular value decomposition (SVD) of the kernel matrix, and the training time of the transductive classifier.

For the baseline GCL methods, the reported training time includes both the encoder training phase and the downstream classifier training. All experiments are conducted using a single 80 GB NVIDIA A100 GPU. The detailed results are provided in Table~\ref{tab:training_time}. As shown, the overall training time of GCL-LRR is comparable to that of state-of-the-art GCL methods such as SFA and MERIT.

\begin{table*}[!htpb]
\small
\centering
\caption{Training time (seconds) comparisons for node classification.}
\label{tab:training_time}
\resizebox{0.95\textwidth}{!}{
\begin{tabular}{|c|cccccccc|}
\hline
Methods  & Cora  & Citeseer & PubMed & Coauthor-CS  &Wiki-CS &Computer &Photo &ogbn-arxiv\\ \hline
GCN      & 11.5  & 13.7  & 38.6  & 43.2   & 22.3  & 30.2  & 19.0  & 215.1 \\
S$^2$GC  & 20.7  & 22.5  & 47.2  & 57.2   & 27.6  & 38.5  & 22.2  & 243.7 \\
GCE      & 32.6  & 36.9  & 67.3  & 80.8   & 37.6  & 50.1  & 32.2  & 346.1 \\
NRGNN    & 72.4  & 80.5  & 142.7 & 189.4  & 74.3  & 97.2  & 62.4  & 650.2 \\
RTGNN    & 143.3 & 169.5 & 299.5 & 353.5  & 153.7 & 201.5 & 124.2 & 1322.2 \\
SUGRL    & 100.3 & 122.1 & 207.4 & 227.1  & 107.7 & 142.8 & 87.7  & 946.8 \\
MERIT    & 167.2 & 179.2 & 336.7 & 375.3  & 172.3 & 226.5 & 140.6 & 1495.1 \\
ARIEL    & 156.9 & 164.3 & 284.3 & 332.6  & 145.1 & 190.4 & 118.3 & 1261.4 \\
SFA      & 237.5 & 269.4 & 457.1 & 492.3  & 233.5 & 304.5 & 187.2 & 2013.1 \\
Sel-Cl   & 177.3 & 189.9 & 313.5 & 352.5  & 161.7 & 211.1 & 130.9 & 1401.1 \\
Jo-SRC   & 148.2 & 157.1 & 281.0 & 306.1  & 144.5 & 188.0 & 118.5 & 1256.0 \\
GRAND+   & 57.4  & 68.4  & 101.7 & 124.2  & 54.8  & 73.8  & 44.5  & 479.2 \\
GCL-LRR   & 159.9 & 174.5 & 350.7 & 380.9  & 180.3 & 235.7 & 145.5 & 1552.7 \\
GCL-LRR   & 166.2 & 185.4 & 372.7 & 399.5  & 195.4 & 253.6 & 159.2 & 1674.8 \\
\hline
\end{tabular}
        }
\end{table*}

\subsection{Ablation Study on the Rank in the Truncated Nuclear Norm}
\label{sec:rank_study}
We conduct an ablation study to investigate the effect of the rank parameter $r_0$ in the TNN $\norm{\bK}{r_0}$ used in the loss function (\ref{eq:loss-GCL-LR-overall}) of GCL-LRR. As shown in Table~\ref{table:noise-rank}, the performance of GCL-LRR remains consistently close to the optimal across different settings of $r_0$, particularly when the rank lies within the range $0.1\min\set{N,d}$ to $0.3\min\set{N,d}$.
\begin{table*}[!htbp]
\centering
\caption{Ablation study on the value of rank $r_0$ in the optimization problem (\ref{eq:node-classification-obj}) on Cora with different levels of asymmetric and symmetric label noise. The accuracy with the optimal rank is shown in the last row. The accuracy difference against the optimal rank is shown for other ranks.}
\label{table:noise-rank}
\resizebox{0.9\textwidth}{!}{

\begin{tabular}{|c|c|cc|cc|cc|}
\hline
\multirow{3}{*}{Rank} & \multicolumn{7}{c|}{Noise Type}                                                    \\
\cmidrule{2-8}
                      & 0      & \multicolumn{2}{c|}{40} & \multicolumn{2}{c|}{60}& \multicolumn{2}{c|}{80} \\
\cmidrule{2-8}
                      & -      & Asymmetric & Symmetric & Asymmetric & Symmetric & Asymmetric & Symmetric  \\
\hline
0.1 $\min\set{N,d}$               & -0.002 & -0.001     &-0.002    & -0.002     & -0.001    & -0.001     & -0.000    \\
0.2 $\min\set{N,d}$              & -0.000 & -0.000     &-0.000    & -0.000     & -0.000    & -0.000     & -0.000    \\
0.3 $\min\set{N,d}$               & -0.000 & -0.000     &-0.001    & -0.002     & -0.001    & -0.000     & -0.001    \\
0.4 $\min\set{N,d}$               & -0.001 & -0.003     &-0.002    & -0.001     & -0.002    & -0.002     & -0.002    \\
0.5 $\min\set{N,d}$               & -0.001 & -0.002     &-0.003    & -0.003     & -0.003    & -0.001     & -0.002    \\
0.6 $\min\set{N,d}$               & -0.003 & -0.002     &-0.002    & -0.003     & -0.002    & -0.002     & -0.003    \\
0.7 $\min\set{N,d}$               & -0.003 & -0.004     &-0.003    & -0.004     & -0.004    & -0.004     & -0.005    \\
0.8 $\min\set{N,d}$               & -0.002 & -0.005     &-0.006    & -0.006     & -0.006    & -0.007     & -0.007    \\
0.9 $\min\set{N,d}$              & -0.004 & -0.004     &-0.005    & -0.007     & -0.008    & -0.008     & -0.006    \\
$\min\set{N,d}$                   & -0.004 & -0.004     &-0.007    & -0.007     & -0.008    & -0.010     & -0.008    \\ \hline
optimal               & 0.858  & 0.589      & 0.713     & 0.492      & 0.587    & 0.306      & 0.419      \\
\hline
\end{tabular}
}
\end{table*}

\subsection{Additional Study in the Kernel Complexity}
\label{sec:sup_kernel_complexity}
In this section, we evaluate the KC of the gram matrix derived from node representations learned by GCL-LRR and other competing GCL methods across various datasets, under symmetric label noise at a corruption level of $40$, as discussed in Theorem~\ref{theorem:optimization-linear-kernel}. The corresponding results are reported in Table~\ref{tab:kernel_complexity}. As observed, the Gram matrices produced by GCL-LRR exhibit substantially lower complexity compared to those of baseline methods. This indicates that transductive classifiers trained on GCL-LRR representations are likely to achieve lower generalization errors on the unlabeled nodes.
\begin{table*}[!htbp]
\centering
\caption{Comparisons in the kernel complexity defined in Theorem~\ref{theorem:optimization-linear-kernel}. The evaluation is performed on the semi-supervised node classification task with $40\%$ of symmetric label noise.}
    \label{tab:kernel_complexity}
\resizebox{0.85\textwidth}{!}{
\begin{tabular}{|c|c|cccccccc|}
\hline
Datasets                          &       & MERIT & SFA  & Jo-SRC & GCN  & GFSA & HONGAT & GCL-LRR & GCL-LR-Attention \\ \hline
\multirow{2}{*}{Cora}             & KC    & 0.37  & 0.42 & 0.48   & 0.44 & 0.35 & 0.40   & 0.20   & 0.18       \\
                                  & $r_0$ & 1420  & 1478 & 1665   & 1511 & 1262 & 1450   & 440    & 395        \\ \hline
\multirow{2}{*}{Citeseer}         & KC    & 0.47  & 0.45 & 0.55   & 0.64 & 0.47 & 0.50   & 0.24   & 0.21       \\
                                  & $r_0$ & 1214  & 1180 & 1405   & 1590 & 1224 & 1285   & 405    & 369        \\ \hline
\multirow{2}{*}{PubMed}           & KC    & 0.54  & 0.50 & 0.62   & 0.71 & 0.52 & 0.66   & 0.30   & 0.28       \\
                                  & $r_0$ & 1644  & 1562 & 1785   & 1993 & 1588 & 1874   & 1197   & 1090       \\ \hline
\multirow{2}{*}{Wiki-CS}          & KC    & 0.42  & 0.44 & 0.40   & 0.49 & 0.43 & 0.45   & 0.19   & 0.17       \\
                                  & $r_0$ & 1805  & 1993 & 1746   & 2130 & 1842 & 2048   & 970    & 904        \\ \hline
\multirow{2}{*}{Amazon-Computers} & KC    & 0.39  & 0.37 & 0.40   & 0.45 & 0.35 & 0.37   & 0.12   & 0.11       \\
                                  & $r_0$ & 1450  & 1428 & 1489   & 1632 & 1370 & 1415   & 874    & 820        \\ \hline
\multirow{2}{*}{Amazon-Photos}    & KC    & 0.38  & 0.38 & 0.43   & 0.47 & 0.39 & 0.41   & 0.14   & 0.12       \\
                                  & $r_0$ & 1872  & 1884 & 1990   & 2145 & 1895 & 1921   & 750    & 722        \\ \hline
\multirow{2}{*}{Coauthor-CS}      & KC    & 0.29  & 0.28 & 0.32   & 0.34 & 0.31 & 0.32   & 0.12   & 0.11       \\
                                  & $r_0$ & 1774  & 1725 & 1896   & 1903 & 1872 & 1890   & 1120   & 1039       \\ \hline
\multirow{2}{*}{ogbn-arxiv}       & KC    & 0.12  & 0.13 & 0.12   & 0.14 & 0.12 & 0.13   & 0.05   & 0.05       \\
                                  & $r_0$ & 1860  & 1936 & 1852   & 1996 & 1845 & 1920   & 1354   & 1328       \\ \hline
\end{tabular}
}
\end{table*}

\subsection{Additional Node Classification Results on More Datasets}
\label{sec:sup_classification-more-data}
Table~\ref{table:label_noise_all_2} presents the node classification results under symmetric label noise, asymmetric label noise, and attribute noise on ogbn-arxiv, Wiki-CS, Amazon-Computers, and Amazon-Photos. The table reports the mean accuracy and standard deviation over $10$ independent runs. As shown, both GCL-LRR and GCL-LR-Attention consistently outperform all baseline methods across these benchmark datasets, demonstrating superior robustness to both label and attribute noise.

\begin{table*}[!htbp]
\centering
\caption{Performance comparison for node classification on ogbn-arxiv, Wiki-CS, Amazon-Computers, and Amazon-Photos with asymmetric label noise, symmetric label noise, and attribute noise.}
\label{table:label_noise_all_2}
\resizebox{\columnwidth}{!}{
\begin{tabular}{|c|c|cccccccccc|}
\hline
  &  &  \multicolumn{10}{c|}{Noise Type}    \\  \cmidrule{3-12}
  &  &  \multicolumn{1}{c|}{0}  &  \multicolumn{3}{c|}{40}  &  \multicolumn{3}{c|}{60}  &  \multicolumn{3}{c|}{80}  \\  \cmidrule{3-12}
\multirow{-3}{*}{Dataset}  &  \multirow{-3}{*}{Methods}  &  \multicolumn{1}{c|}{-}  &  Asymmetric  &  Symmetric  &  \multicolumn{1}{c|}{Attribute}  &  Asymmetric  &  Symmetric  &  \multicolumn{1}{c|}{Attribute}  &  Asymmetric  &  Symmetric  &  Attribute  \\  \hline
\multirow{19}{*}{ogbn-arxiv}  &  GCN  &  \multicolumn{1}{c|}{0.717$\pm$0.003}  &  0.401$\pm$0.014  &  0.421$\pm$0.014  &  \multicolumn{1}{c|}{0.478$\pm$0.010}  &  0.336$\pm$0.011  &  0.346$\pm$0.021  &  \multicolumn{1}{c|}{0.339$\pm$0.012}  &  0.286$\pm$0.022  &  0.256$\pm$0.010  &  0.294$\pm$0.013  \\
  &  S$^2$GC  &  \multicolumn{1}{c|}{0.712$\pm$0.003}  &  0.417$\pm$0.017  &  0.429$\pm$0.014  &  \multicolumn{1}{c|}{0.492$\pm$0.010}  &  0.344$\pm$0.016  &  0.353$\pm$0.031  &  \multicolumn{1}{c|}{0.343$\pm$0.009}  &  0.297$\pm$0.023  &  0.266$\pm$0.013  &  0.284$\pm$0.012  \\
  &  GCE  &  \multicolumn{1}{c|}{0.720$\pm$0.004}  &  0.410$\pm$0.018  &  0.428$\pm$0.008  &  \multicolumn{1}{c|}{0.480$\pm$0.014}  &  0.348$\pm$0.019  &  0.344$\pm$0.019  &  \multicolumn{1}{c|}{0.342$\pm$0.015}  &  0.310$\pm$0.014  &  0.260$\pm$0.011  &  0.275$\pm$0.015  \\
  &  UnionNET  &  \multicolumn{1}{c|}{0.724$\pm$0.006}  &  0.429$\pm$0.021  &  0.449$\pm$0.007  &  \multicolumn{1}{c|}{0.485$\pm$0.012}  &  0.362$\pm$0.018  &  0.367$\pm$0.008  &  \multicolumn{1}{c|}{0.340$\pm$0.009}  &  0.332$\pm$0.019  &  0.269$\pm$0.013  &  0.280$\pm$0.012  \\
  &  NRGNN  &  \multicolumn{1}{c|}{0.721$\pm$0.006}  &  0.449$\pm$0.014  &  0.466$\pm$0.009  &  \multicolumn{1}{c|}{0.485$\pm$0.012}  &  0.371$\pm$0.020  &  0.379$\pm$0.008  &  \multicolumn{1}{c|}{0.342$\pm$0.011}  &  0.330$\pm$0.018  &  0.271$\pm$0.018  &  0.300$\pm$0.010  \\
  &  RTGNN  &  \multicolumn{1}{c|}{0.718$\pm$0.004}  &  0.443$\pm$0.012  &  0.464$\pm$0.012  &  \multicolumn{1}{c|}{0.484$\pm$0.014}  &  0.380$\pm$0.011  &  0.384$\pm$0.013  &  \multicolumn{1}{c|}{0.340$\pm$0.017}  &  0.335$\pm$0.011  &  0.285$\pm$0.015  &  0.301$\pm$0.006  \\
  &  SUGRL  &  \multicolumn{1}{c|}{0.693$\pm$0.002}  &  0.439$\pm$0.010  &  0.467$\pm$0.010  &  \multicolumn{1}{c|}{0.480$\pm$0.012}  &  0.365$\pm$0.013  &  0.385$\pm$0.011  &  \multicolumn{1}{c|}{0.341$\pm$0.009}  &  0.327$\pm$0.011  &  0.275$\pm$0.011  &  0.295$\pm$0.011  \\
  &  MERIT  &  \multicolumn{1}{c|}{0.717$\pm$0.004}  &  0.442$\pm$0.009  &  0.463$\pm$0.009  &  \multicolumn{1}{c|}{0.483$\pm$0.010}  &  0.368$\pm$0.011  &  0.381$\pm$0.011  &  \multicolumn{1}{c|}{0.341$\pm$0.012}  &  0.324$\pm$0.012  &  0.272$\pm$0.010  &  0.304$\pm$0.009  \\
  &  ARIEL  &  \multicolumn{1}{c|}{0.717$\pm$0.004}  &  0.448$\pm$0.013  &  0.471$\pm$0.013  &  \multicolumn{1}{c|}{0.482$\pm$0.011}  &  0.379$\pm$0.014  &  0.384$\pm$0.015  &  \multicolumn{1}{c|}{0.342$\pm$0.015}  &  0.334$\pm$0.014  &  0.280$\pm$0.013  &  0.300$\pm$0.010  \\
  &  SFA  &  \multicolumn{1}{c|}{0.718$\pm$0.009}&  0.445$\pm$0.012  &  0.463$\pm$0.013  &  \multicolumn{1}{c|}{0.486$\pm$0.012}  &  0.368$\pm$0.011  &  0.378$\pm$0.014  &  \multicolumn{1}{c|}{0.338$\pm$0.015}  &  0.325$\pm$0.014  &  0.273$\pm$0.012  &  0.302$\pm$0.013  \\
  &  Sel-Cl  &  \multicolumn{1}{c|}{0.719$\pm$0.002}  &  0.447$\pm$0.007  &  0.469$\pm$0.007  &  \multicolumn{1}{c|}{0.486$\pm$0.010}  &  0.375$\pm$0.008  &  0.389$\pm$0.025  &  \multicolumn{1}{c|}{0.344$\pm$0.013}  &  0.331$\pm$0.008  &  0.284$\pm$0.019  &  0.304$\pm$0.012  \\
  &  Jo-SRC  &  \multicolumn{1}{c|}{0.715$\pm$0.005}  &  0.445$\pm$0.011  &  0.466$\pm$0.009  &  \multicolumn{1}{c|}{0.481$\pm$0.010}  &  0.377$\pm$0.013  &  0.387$\pm$0.013  &  \multicolumn{1}{c|}{0.340$\pm$0.013}  &  0.333$\pm$0.013  &  0.282$\pm$0.018  &  0.297$\pm$0.009  \\
  &  GRAND+  &  \multicolumn{1}{c|}{0.725$\pm$0.004}  &  0.445$\pm$0.008  &  0.466$\pm$0.011  &  \multicolumn{1}{c|}{0.481$\pm$0.011}  &  0.378$\pm$0.010  &  0.385$\pm$0.012  &  \multicolumn{1}{c|}{0.344$\pm$0.010}  &  0.332$\pm$0.010  &  0.282$\pm$0.016  &  0.303$\pm$0.009  \\
  &  GFSA  &  \multicolumn{1}{c|}{0.719$\pm$0.004}  &  0.443$\pm$0.012  &  0.460$\pm$0.010  &  \multicolumn{1}{c|}{0.482$\pm$0.011}  &  0.370$\pm$0.012  &  0.379$\pm$0.012  &  \multicolumn{1}{c|}{0.342$\pm$0.011}  &  0.328$\pm$0.012  &  0.278$\pm$0.013  &  0.299$\pm$0.011  \\
  &  HONGAT  &  \multicolumn{1}{c|}{0.716$\pm$0.005}  &  0.440$\pm$0.011  &  0.458$\pm$0.012  &  \multicolumn{1}{c|}{0.480$\pm$0.012}  &  0.366$\pm$0.013  &  0.373$\pm$0.013  &  \multicolumn{1}{c|}{0.339$\pm$0.012}  &  0.324$\pm$0.014  &  0.276$\pm$0.014  &  0.296$\pm$0.012  \\
  &  CRGNN  &  \multicolumn{1}{c|}{0.721$\pm$0.003}  &  0.446$\pm$0.010  &  0.465$\pm$0.010  &  \multicolumn{1}{c|}{0.483$\pm$0.009}  &  0.372$\pm$0.010  &  0.382$\pm$0.011  &  \multicolumn{1}{c|}{0.343$\pm$0.010}  &  0.330$\pm$0.012  &  0.281$\pm$0.012  &  0.302$\pm$0.010  \\
  &  CGNN  &  \multicolumn{1}{c|}{0.717$\pm$0.006}  &  0.441$\pm$0.013  &  0.462$\pm$0.011  &  \multicolumn{1}{c|}{0.481$\pm$0.010}  &  0.368$\pm$0.014  &  0.376$\pm$0.012  &  \multicolumn{1}{c|}{0.340$\pm$0.011}  &  0.326$\pm$0.015  &  0.277$\pm$0.013  &  0.298$\pm$0.012  \\
  &  GCL-LRR  &  \multicolumn{1}{c|}{\underline{0.728$\pm$0.006}}  &  \underline{0.472$\pm$0.013}  &  \underline{0.492$\pm$0.011}  &  \multicolumn{1}{c|}{\underline{0.508$\pm$0.014}}  & \underline{0.405$\pm$0.014}  &  \underline{0.411$\pm$0.012}  &  \multicolumn{1}{c|}{\underline{0.405$\pm$0.012}}  & \underline{0.359$\pm$0.015}  &  \underline{0.307$\pm$0.013}  &  \underline{0.335$\pm$0.013}  \\
  & GCL-LR-Attention & \multicolumn{1}{c|}{\textbf{0.731}$\pm$\textbf{0.006}} & \textbf{0.487}$\pm$\textbf{0.013} & \textbf{0.507}$\pm$\textbf{0.011} & \multicolumn{1}{c|}{\textbf{0.523}$\pm$\textbf{0.014}} & \textbf{0.423}$\pm$\textbf{0.014} & \textbf{0.430}$\pm$\textbf{0.012} & \multicolumn{1}{c|}{\textbf{0.423}$\pm$\textbf{0.012}} & \textbf{0.374}$\pm$\textbf{0.015} & \textbf{0.332}$\pm$\textbf{0.013} & \textbf{0.350}$\pm$\textbf{0.013} \\ \hline\multirow{19}{*}{Wiki-CS}  &  GCN  &  \multicolumn{1}{c|}{0.918$\pm$0.001}  &  0.645$\pm$0.009  &  0.656$\pm$0.006  &  \multicolumn{1}{c|}{0.702$\pm$0.010}  &  0.511$\pm$0.013  &  0.501$\pm$0.009  &  \multicolumn{1}{c|}{0.531$\pm$0.010}  &  0.429$\pm$0.022  &  0.389$\pm$0.011  &  0.415$\pm$0.013  \\
  &  S$^2$GC  &  \multicolumn{1}{c|}{0.918$\pm$0.001}  &  0.657$\pm$0.012  &  0.663$\pm$0.006  &  \multicolumn{1}{c|}{0.713$\pm$0.010}  &  0.516$\pm$0.013  &  0.514$\pm$0.009  &  \multicolumn{1}{c|}{0.556$\pm$0.009}  &  0.437$\pm$0.020  &  0.396$\pm$0.010  &  0.422$\pm$0.012  \\
  &  GCE  &  \multicolumn{1}{c|}{0.922$\pm$0.003}  &  0.662$\pm$0.017  &  0.659$\pm$0.007  &  \multicolumn{1}{c|}{0.705$\pm$0.014}  &  0.515$\pm$0.016  &  0.502$\pm$0.007  &  \multicolumn{1}{c|}{0.539$\pm$0.009}  &  0.443$\pm$0.017  &  0.389$\pm$0.012  &  0.412$\pm$0.011  \\
  &  UnionNET  &  \multicolumn{1}{c|}{0.918$\pm$0.002}  &  0.669$\pm$0.023  &  0.671$\pm$0.013  &  \multicolumn{1}{c|}{0.706$\pm$0.012}  &  0.525$\pm$0.011  &  0.529$\pm$0.011  &  \multicolumn{1}{c|}{0.540$\pm$0.012}  &  0.458$\pm$0.015  &  0.401$\pm$0.011  &  0.420$\pm$0.007  \\
  &  NRGNN  &  \multicolumn{1}{c|}{0.919$\pm$0.002}  &  0.678$\pm$0.014  &  0.689$\pm$0.009  &  \multicolumn{1}{c|}{0.705$\pm$0.012}  &  0.545$\pm$0.021  &  0.556$\pm$0.011  &  \multicolumn{1}{c|}{0.546$\pm$0.011}  &  0.461$\pm$0.012  &  0.410$\pm$0.012  &  0.417$\pm$0.007  \\
  &  RTGNN  &  \multicolumn{1}{c|}{0.920$\pm$0.005}  &  0.678$\pm$0.012  &  0.691$\pm$0.009  &  \multicolumn{1}{c|}{0.712$\pm$0.008}  &  0.559$\pm$0.010  &  0.569$\pm$0.011  &  \multicolumn{1}{c|}{0.560$\pm$0.008}  &  0.455$\pm$0.015  &  0.415$\pm$0.015  &  0.412$\pm$0.014  \\
  &  SUGRL  &  \multicolumn{1}{c|}{0.922$\pm$0.005}  &  0.675$\pm$0.010  &  0.695$\pm$0.010  &  \multicolumn{1}{c|}{0.714$\pm$0.006}  &  0.550$\pm$0.011  &  0.560$\pm$0.011  &  \multicolumn{1}{c|}{0.561$\pm$0.007}  &  0.449$\pm$0.011  &  0.411$\pm$0.011  &  0.429$\pm$0.008  \\
  &  MERIT  &  \multicolumn{1}{c|}{0.924$\pm$0.004}  &  0.679$\pm$0.011  &  0.689$\pm$0.008  &  \multicolumn{1}{c|}{0.709$\pm$0.005}  &  0.552$\pm$0.014  &  0.562$\pm$0.014  &  \multicolumn{1}{c|}{0.562$\pm$0.011}  &  0.452$\pm$0.013  &  0.403$\pm$0.013  &  0.426$\pm$0.005  \\
  &  ARIEL  &  \multicolumn{1}{c|}{0.925$\pm$0.004}  &  0.682$\pm$0.011  &  0.699$\pm$0.009  &  \multicolumn{1}{c|}{0.712$\pm$0.005}  &  0.555$\pm$0.011  &  0.566$\pm$0.011  &  \multicolumn{1}{c|}{0.556$\pm$0.011}  &  0.454$\pm$0.014  &  0.415$\pm$0.019  &  0.427$\pm$0.013  \\
  &  SFA  &  \multicolumn{1}{c|}{0.925$\pm$0.009}&  0.682$\pm$0.011  &  0.690$\pm$0.012  &  \multicolumn{1}{c|}{0.715$\pm$0.012}  &  0.555$\pm$0.015  &  0.567$\pm$0.014  &  \multicolumn{1}{c|}{0.565$\pm$0.013}  &  0.458$\pm$0.013  &  0.402$\pm$0.013  &  0.429$\pm$0.015  \\
  &  Sel-Cl  &  \multicolumn{1}{c|}{0.922$\pm$0.008}  &  0.684$\pm$0.009  &  0.694$\pm$0.012  &  \multicolumn{1}{c|}{0.714$\pm$0.010}  &  0.557$\pm$0.013  &  0.568$\pm$0.013  &  \multicolumn{1}{c|}{0.566$\pm$0.010}  &  0.457$\pm$0.013  &  0.412$\pm$0.017  &  0.425$\pm$0.009  \\
  &  Jo-SRC  &  \multicolumn{1}{c|}{0.921$\pm$0.005}  &  0.684$\pm$0.011  &  0.695$\pm$0.004  &  \multicolumn{1}{c|}{0.709$\pm$0.007}  &  0.560$\pm$0.011  &  0.566$\pm$0.011  &  \multicolumn{1}{c|}{0.561$\pm$0.009}  &  0.456$\pm$0.013  &  0.410$\pm$0.018  &  0.428$\pm$0.010  \\
  &  GRAND+  &  \multicolumn{1}{c|}{0.927$\pm$0.004}  &  0.682$\pm$0.011  &  0.693$\pm$0.006  &  \multicolumn{1}{c|}{0.715$\pm$0.008}  &  0.554$\pm$0.008  &  0.568$\pm$0.013  &  \multicolumn{1}{c|}{0.557$\pm$0.011}  &  0.455$\pm$0.012  &  0.416$\pm$0.013  &  0.428$\pm$0.011  \\
  &  GFSA  &  \multicolumn{1}{c|}{0.923$\pm$0.004}  &  0.680$\pm$0.012  &  0.691$\pm$0.008  &  \multicolumn{1}{c|}{0.711$\pm$0.010}  &  0.553$\pm$0.010  &  0.562$\pm$0.011  &  \multicolumn{1}{c|}{0.560$\pm$0.010}  &  0.453$\pm$0.014  &  0.408$\pm$0.012  &  0.423$\pm$0.010  \\
  &  HONGAT  &  \multicolumn{1}{c|}{0.921$\pm$0.003}  &  0.674$\pm$0.014  &  0.685$\pm$0.010  &  \multicolumn{1}{c|}{0.707$\pm$0.011}  &  0.546$\pm$0.012  &  0.553$\pm$0.010  &  \multicolumn{1}{c|}{0.552$\pm$0.010}  &  0.448$\pm$0.014  &  0.404$\pm$0.013  &  0.419$\pm$0.012  \\
  &  CRGNN  &  \multicolumn{1}{c|}{0.924$\pm$0.005}  &  0.683$\pm$0.011  &  0.696$\pm$0.008  &  \multicolumn{1}{c|}{0.713$\pm$0.008}  &  0.557$\pm$0.010  &  0.565$\pm$0.012  &  \multicolumn{1}{c|}{0.564$\pm$0.009}  &  0.456$\pm$0.013  &  0.411$\pm$0.012  &  0.426$\pm$0.010  \\
  &  CGNN  &  \multicolumn{1}{c|}{0.920$\pm$0.004}  &  0.677$\pm$0.010  &  0.688$\pm$0.009  &  \multicolumn{1}{c|}{0.710$\pm$0.011}  &  0.549$\pm$0.011  &  0.559$\pm$0.013  &  \multicolumn{1}{c|}{0.558$\pm$0.010}  &  0.451$\pm$0.015  &  0.406$\pm$0.012  &  0.421$\pm$0.009  \\
  &  GCL-LRR  &  \multicolumn{1}{c|}{\underline{0.933$\pm$0.006}}  &  \underline{0.699$\pm$0.015}  &  \underline{0.721$\pm$0.011}  &  \multicolumn{1}{c|}{\underline{0.742$\pm$0.015}}  & \underline{0.575$\pm$0.014}  &  \underline{0.595$\pm$0.018}  &  \multicolumn{1}{c|}{\underline{0.588$\pm$0.015}}  &  \underline{0.469$\pm$0.015}  &  \underline{0.438$\pm$0.015}  &  \underline{0.453$\pm$0.017}  \\
  & GCL-LR-Attention & \multicolumn{1}{c|}{\textbf{0.936}$\pm$\textbf{0.006}} & \textbf{0.714}$\pm$\textbf{0.015} & \textbf{0.736}$\pm$\textbf{0.011} & \multicolumn{1}{c|}{\textbf{0.758}$\pm$\textbf{0.015}} & \textbf{0.594}$\pm$\textbf{0.014} & \textbf{0.612}$\pm$\textbf{0.018} & \multicolumn{1}{c|}{\textbf{0.606}$\pm$\textbf{0.015}} & \textbf{0.489}$\pm$\textbf{0.015} & \textbf{0.453}$\pm$\textbf{0.015} & \textbf{0.470}$\pm$\textbf{0.017} \\ \hline
\multirow{19}{*}{Amazon-Computers}  &  GCN   &  \multicolumn{1}{c|}{0.815$\pm$0.005}  &  0.547$\pm$0.015  &  0.636$\pm$0.007  &  \multicolumn{1}{c|}{0.639$\pm$0.008}  &  0.405$\pm$0.014  &  0.517$\pm$0.010  &  \multicolumn{1}{c|}{0.439$\pm$0.012}  &  0.265$\pm$0.012  &  0.354$\pm$0.014  &  0.317$\pm$0.013  \\
  &  S$^2$GC  &  \multicolumn{1}{c|}{0.835$\pm$0.002}  &  0.569$\pm$0.007  &  0.664$\pm$0.007  &  \multicolumn{1}{c|}{0.661$\pm$0.007}  &  0.422$\pm$0.010  &  0.535$\pm$0.010  &  \multicolumn{1}{c|}{0.454$\pm$0.011}  &  0.279$\pm$0.014  &  0.366$\pm$0.014  &  0.320$\pm$0.013  \\
  &  GCE  &  \multicolumn{1}{c|}{0.819$\pm$0.004}  &  0.573$\pm$0.011  &  0.652$\pm$0.008  &  \multicolumn{1}{c|}{0.650$\pm$0.014}  &  0.449$\pm$0.011  &  0.509$\pm$0.011  &  \multicolumn{1}{c|}{0.445$\pm$0.015}  &  0.280$\pm$0.013  &  0.353$\pm$0.013  &  0.325$\pm$0.015  \\
  &  UnionNET  &  \multicolumn{1}{c|}{0.820$\pm$0.006}  &  0.569$\pm$0.014  &  0.664$\pm$0.007  &  \multicolumn{1}{c|}{0.653$\pm$0.012}  &  0.452$\pm$0.010  &  0.541$\pm$0.010  &  \multicolumn{1}{c|}{0.450$\pm$0.009}  &  0.283$\pm$0.014  &  0.370$\pm$0.011  &  0.320$\pm$0.012  \\
  &  NRGNN  &  \multicolumn{1}{c|}{0.822$\pm$0.006}  &  0.571$\pm$0.019  &  0.676$\pm$0.007  &  \multicolumn{1}{c|}{0.645$\pm$0.012}  &  0.470$\pm$0.014  &  0.548$\pm$0.014  &  \multicolumn{1}{c|}{0.451$\pm$0.011}  &  0.282$\pm$0.022  &  0.373$\pm$0.012  &  0.326$\pm$0.010  \\
  &  RTGNN  &  \multicolumn{1}{c|}{0.828$\pm$0.003}  &  0.570$\pm$0.010  &  0.682$\pm$0.008  &  \multicolumn{1}{c|}{0.678$\pm$0.011}  &  0.474$\pm$0.011  &  0.555$\pm$0.010  &  \multicolumn{1}{c|}{0.457$\pm$0.009}  &  0.280$\pm$0.011  &  0.386$\pm$0.014  &  0.342$\pm$0.016  \\
  &  SUGRL  &  \multicolumn{1}{c|}{0.834$\pm$0.005}  &  0.564$\pm$0.011  &  0.674$\pm$0.012  &  \multicolumn{1}{c|}{0.675$\pm$0.009}  &  0.468$\pm$0.011  &  0.552$\pm$0.011  &  \multicolumn{1}{c|}{0.452$\pm$0.012}  &  0.280$\pm$0.012  &  0.381$\pm$0.012  &  0.338$\pm$0.014  \\
  &  MERIT  &  \multicolumn{1}{c|}{0.831$\pm$0.005}  &  0.560$\pm$0.008  &  0.670$\pm$0.008  &  \multicolumn{1}{c|}{0.671$\pm$0.009}  &  0.467$\pm$0.013  &  0.547$\pm$0.013  &  \multicolumn{1}{c|}{0.450$\pm$0.014}  &  0.277$\pm$0.013  &  0.385$\pm$0.013  &  0.335$\pm$0.009  \\
  &  ARIEL  &  \multicolumn{1}{c|}{0.843$\pm$0.004}  &  0.573$\pm$0.013  &  0.681$\pm$0.010  &  \multicolumn{1}{c|}{0.675$\pm$0.009}  &  0.471$\pm$0.012  &  0.553$\pm$0.012  &  \multicolumn{1}{c|}{0.455$\pm$0.014}  &  0.284$\pm$0.014  &  0.389$\pm$0.013  &  0.343$\pm$0.013  \\
  &  SFA  &  \multicolumn{1}{c|}{0.839$\pm$0.010}  &  0.564$\pm$0.011  &  0.677$\pm$0.013  &  \multicolumn{1}{c|}{0.676$\pm$0.015}  &  0.473$\pm$0.014  &  0.549$\pm$0.014  &  \multicolumn{1}{c|}{0.457$\pm$0.014}  &  0.282$\pm$0.016  &  0.389$\pm$0.013  &  0.344$\pm$0.017  \\
  &  Sel-Cl  &  \multicolumn{1}{c|}{0.828$\pm$0.002}  &  0.570$\pm$0.010  &  0.685$\pm$0.012  &  \multicolumn{1}{c|}{0.676$\pm$0.009}  &  0.472$\pm$0.013  &  0.554$\pm$0.014  &  \multicolumn{1}{c|}{0.455$\pm$0.011}  &  0.282$\pm$0.017  &  0.389$\pm$0.013  &  0.341$\pm$0.015  \\
  &  Jo-SRC  &  \multicolumn{1}{c|}{0.825$\pm$0.005}  &  0.571$\pm$0.006  &  0.684$\pm$0.013  &  \multicolumn{1}{c|}{0.679$\pm$0.007}  &  0.473$\pm$0.011  &  0.556$\pm$0.008  &  \multicolumn{1}{c|}{0.458$\pm$0.012}  &  0.285$\pm$0.013  &  0.387$\pm$0.018  &  0.345$\pm$0.018  \\
  &  GRAND+  &  \multicolumn{1}{c|}{0.858$\pm$0.006}  &  0.570$\pm$0.009  &  0.682$\pm$0.007  &  \multicolumn{1}{c|}{0.678$\pm$0.011}  &  0.472$\pm$0.010  &  0.554$\pm$0.008  &  \multicolumn{1}{c|}{0.456$\pm$0.012}  &  0.284$\pm$0.015  &  0.387$\pm$0.015  &  0.345$\pm$0.013  \\
  &  GFSA  &  \multicolumn{1}{c|}{0.837$\pm$0.004}  &  0.567$\pm$0.010  &  0.672$\pm$0.009  &  \multicolumn{1}{c|}{0.667$\pm$0.010}  &  0.463$\pm$0.012  &  0.543$\pm$0.011  &  \multicolumn{1}{c|}{0.453$\pm$0.012}  &  0.281$\pm$0.014  &  0.376$\pm$0.013  &  0.333$\pm$0.014  \\
  &  HONGAT  &  \multicolumn{1}{c|}{0.841$\pm$0.005}  &  0.571$\pm$0.008  &  0.678$\pm$0.011  &  \multicolumn{1}{c|}{0.673$\pm$0.012}  &  0.469$\pm$0.013  &  0.551$\pm$0.012  &  \multicolumn{1}{c|}{0.456$\pm$0.011}  &  0.283$\pm$0.015  &  0.384$\pm$0.014  &  0.340$\pm$0.015  \\
  &  CRGNN  &  \multicolumn{1}{c|}{0.846$\pm$0.003}  &  0.572$\pm$0.009  &  0.680$\pm$0.008  &  \multicolumn{1}{c|}{0.677$\pm$0.009}  &  0.471$\pm$0.011  &  0.553$\pm$0.010  &  \multicolumn{1}{c|}{0.457$\pm$0.010}  &  0.284$\pm$0.013  &  0.388$\pm$0.012  &  0.342$\pm$0.012  \\
  &  CGNN  &  \multicolumn{1}{c|}{0.844$\pm$0.004}  &  0.569$\pm$0.011  &  0.675$\pm$0.010  &  \multicolumn{1}{c|}{0.670$\pm$0.011}  &  0.466$\pm$0.012  &  0.548$\pm$0.011  &  \multicolumn{1}{c|}{0.454$\pm$0.013}  &  0.282$\pm$0.014  &  0.382$\pm$0.013  &  0.337$\pm$0.014  \\
  &  GCL-LRR  &  \multicolumn{1}{c|}{\underline{0.858$\pm$0.006}}  &  \underline{0.589$\pm$0.011}  &  \underline{0.713$\pm$0.007}  &  \multicolumn{1}{c|}{\underline{0.695$\pm$0.011}}  & \underline{0.492$\pm$0.011}  &  \underline{0.587$\pm$0.013}  &  \multicolumn{1}{c|}{\underline{0.477$\pm$0.012}}  &  \underline{0.306$\pm$0.012}  &  \underline{0.419$\pm$0.012}  & \underline{0.363$\pm$0.011}  \\
  & GCL-LR-Attention & \multicolumn{1}{c|}{\textbf{0.861}$\pm$\textbf{0.006}} & \textbf{0.602}$\pm$\textbf{0.011} & \textbf{0.724}$\pm$\textbf{0.007} & \multicolumn{1}{c|}{\textbf{0.708}$\pm$\textbf{0.011}} & \textbf{0.510}$\pm$\textbf{0.011} & \textbf{0.605}$\pm$\textbf{0.013} & \multicolumn{1}{c|}{\textbf{0.492}$\pm$\textbf{0.012}} & \textbf{0.329}$\pm$\textbf{0.012} & \textbf{0.436}$\pm$\textbf{0.012} & \textbf{0.382}$\pm$\textbf{0.011} \\ \hline
\multirow{19}{*}{Amazon-Photos}  &  GCN  &  \multicolumn{1}{c|}{0.703$\pm$0.005}  &  0.475$\pm$0.023  &  0.501$\pm$0.013  &  \multicolumn{1}{c|}{0.529$\pm$0.009}  &  0.351$\pm$0.014  &  0.341$\pm$0.014  &  \multicolumn{1}{c|}{0.372$\pm$0.011}  &  0.291$\pm$0.022  &  0.281$\pm$0.019  &  0.290$\pm$0.014  \\
  &  S$^2$GC  &  \multicolumn{1}{c|}{0.736$\pm$0.005}  &  0.488$\pm$0.013  &  0.528$\pm$0.013  &  \multicolumn{1}{c|}{0.553$\pm$0.008}  &  0.363$\pm$0.012  &  0.367$\pm$0.014  &  \multicolumn{1}{c|}{0.390$\pm$0.013}  &  0.304$\pm$0.024  &  0.284$\pm$0.019  &  0.288$\pm$0.011  \\
  &  GCE  &  \multicolumn{1}{c|}{0.705$\pm$0.004}  &  0.490$\pm$0.016  &  0.512$\pm$0.014  &  \multicolumn{1}{c|}{0.540$\pm$0.014}  &  0.362$\pm$0.015  &  0.352$\pm$0.010  &  \multicolumn{1}{c|}{0.381$\pm$0.009}  &  0.309$\pm$0.012  &  0.285$\pm$0.014  &  0.285$\pm$0.011  \\
  &  UnionNET  &  \multicolumn{1}{c|}{0.706$\pm$0.006}  &  0.499$\pm$0.015  &  0.547$\pm$0.014  &  \multicolumn{1}{c|}{0.545$\pm$0.013}  &  0.379$\pm$0.013  &  0.399$\pm$0.013  &  \multicolumn{1}{c|}{0.379$\pm$0.012}  &  0.322$\pm$0.021  &  0.302$\pm$0.013  &  0.290$\pm$0.012  \\
  &  NRGNN  &  \multicolumn{1}{c|}{0.710$\pm$0.006}  &  0.498$\pm$0.015  &  0.546$\pm$0.015  &  \multicolumn{1}{c|}{0.538$\pm$0.011}  &  0.382$\pm$0.016  &  0.412$\pm$0.016  &  \multicolumn{1}{c|}{0.377$\pm$0.012}  &  0.336$\pm$0.021  &  0.309$\pm$0.018  &  0.284$\pm$0.009  \\
  &  RTGNN  &  \multicolumn{1}{c|}{0.746$\pm$0.008}  &  0.498$\pm$0.007  &  0.556$\pm$0.007  &  \multicolumn{1}{c|}{0.550$\pm$0.012}  &  0.392$\pm$0.010  &  0.424$\pm$0.013  &  \multicolumn{1}{c|}{0.390$\pm$0.014}  &  0.348$\pm$0.017  &  0.308$\pm$0.016  &  0.302$\pm$0.011  \\
  &  SUGRL  &  \multicolumn{1}{c|}{0.730$\pm$0.005}  &  0.493$\pm$0.011  &  0.541$\pm$0.011  &  \multicolumn{1}{c|}{0.544$\pm$0.010}  &  0.376$\pm$0.009  &  0.421$\pm$0.009  &  \multicolumn{1}{c|}{0.388$\pm$0.009}  &  0.339$\pm$0.010  &  0.305$\pm$0.010  &  0.300$\pm$0.009  \\
  &  MERIT  &  \multicolumn{1}{c|}{0.740$\pm$0.007}  &  0.496$\pm$0.012  &  0.536$\pm$0.012  &  \multicolumn{1}{c|}{0.542$\pm$0.010}  &  0.383$\pm$0.011  &  0.425$\pm$0.011  &  \multicolumn{1}{c|}{0.387$\pm$0.008}  &  0.344$\pm$0.014  &  0.301$\pm$0.014  &  0.295$\pm$0.009  \\
  &  SFA  &  \multicolumn{1}{c|}{0.740$\pm$0.011}  &  0.502$\pm$0.014  &  0.532$\pm$0.015  &  \multicolumn{1}{c|}{0.547$\pm$0.013}  &  0.390$\pm$0.014  &  0.433$\pm$0.014  &  \multicolumn{1}{c|}{0.389$\pm$0.012}  &  0.347$\pm$0.016  &  0.312$\pm$0.015  &  0.299$\pm$0.013  \\
  &  ARIEL  &  \multicolumn{1}{c|}{0.727$\pm$0.007}  &  0.500$\pm$0.008  &  0.550$\pm$0.013  &  \multicolumn{1}{c|}{0.548$\pm$0.008}  &  0.391$\pm$0.009  &  0.427$\pm$0.012  &  \multicolumn{1}{c|}{0.389$\pm$0.014}  &  0.349$\pm$0.014  &  0.307$\pm$0.013  &  0.299$\pm$0.013  \\
  &  Sel-Cl  &  \multicolumn{1}{c|}{0.725$\pm$0.008}  &  0.499$\pm$0.012  &  0.551$\pm$0.010  &  \multicolumn{1}{c|}{0.549$\pm$0.008}  &  0.389$\pm$0.011  &  0.426$\pm$0.008  &  \multicolumn{1}{c|}{0.391$\pm$0.020}  &  0.350$\pm$0.018  &  0.310$\pm$0.015  &  0.300$\pm$0.017  \\
  &  Jo-SRC  &  \multicolumn{1}{c|}{0.730$\pm$0.005}  &  0.500$\pm$0.013  &  0.555$\pm$0.011  &  \multicolumn{1}{c|}{0.551$\pm$0.011}  &  0.394$\pm$0.013  &  0.425$\pm$0.013  &  \multicolumn{1}{c|}{0.393$\pm$0.013}  &  0.351$\pm$0.013  &  0.305$\pm$0.018  &  0.303$\pm$0.013  \\
  &  GRAND+  &  \multicolumn{1}{c|}{0.756$\pm$0.004}  &  0.497$\pm$0.010  &  0.553$\pm$0.010  &  \multicolumn{1}{c|}{0.552$\pm$0.011}  &  0.390$\pm$0.013  &  0.422$\pm$0.013  &  \multicolumn{1}{c|}{0.387$\pm$0.013}  &  0.348$\pm$0.013  &  0.309$\pm$0.014  &  0.302$\pm$0.012  \\
  &  GFSA  &  \multicolumn{1}{c|}{0.722$\pm$0.006}  &  0.492$\pm$0.012  &  0.530$\pm$0.012  &  \multicolumn{1}{c|}{0.543$\pm$0.010}  &  0.372$\pm$0.012  &  0.398$\pm$0.011  &  \multicolumn{1}{c|}{0.383$\pm$0.011}  &  0.328$\pm$0.016  &  0.294$\pm$0.015  &  0.292$\pm$0.012  \\
  &  HONGAT  &  \multicolumn{1}{c|}{0.738$\pm$0.005}  &  0.496$\pm$0.010  &  0.542$\pm$0.011  &  \multicolumn{1}{c|}{0.547$\pm$0.009}  &  0.384$\pm$0.013  &  0.415$\pm$0.012  &  \multicolumn{1}{c|}{0.388$\pm$0.012}  &  0.342$\pm$0.015  &  0.303$\pm$0.014  &  0.298$\pm$0.011  \\
  &  CRGNN  &  \multicolumn{1}{c|}{0.744$\pm$0.004}  &  0.501$\pm$0.009  &  0.548$\pm$0.010  &  \multicolumn{1}{c|}{0.549$\pm$0.008}  &  0.388$\pm$0.011  &  0.422$\pm$0.010  &  \multicolumn{1}{c|}{0.390$\pm$0.010}  &  0.346$\pm$0.014  &  0.306$\pm$0.013  &  0.301$\pm$0.010  \\
  &  CGNN  &  \multicolumn{1}{c|}{0.732$\pm$0.007}  &  0.494$\pm$0.011  &  0.538$\pm$0.013  &  \multicolumn{1}{c|}{0.545$\pm$0.011}  &  0.378$\pm$0.012  &  0.408$\pm$0.013  &  \multicolumn{1}{c|}{0.385$\pm$0.013}  &  0.335$\pm$0.017  &  0.300$\pm$0.016  &  0.296$\pm$0.013  \\
  &  GCL-LRR  &  \multicolumn{1}{c|}{\underline{0.757$\pm$0.010}}  &  \underline{0.520$\pm$0.013}  &  \underline{0.581$\pm$0.013}  &  \multicolumn{1}{c|}{\underline{0.570$\pm$0.007}}  &  \underline{0.410$\pm$0.014}  &  \underline{0.455$\pm$0.014}  &  \multicolumn{1}{c|}{\underline{0.406$\pm$0.012}}  &  \underline{0.369$\pm$0.012}  &  \underline{0.335$\pm$0.014}  &  \underline{0.318$\pm$0.010}  \\
  & GCL-LR-Attention & \multicolumn{1}{c|}{\textbf{0.762}$\pm$\textbf{0.010}} & \textbf{0.533}$\pm$\textbf{0.013} & \textbf{0.597}$\pm$\textbf{0.013} & \multicolumn{1}{c|}{\textbf{0.588}$\pm$\textbf{0.007}} & \textbf{0.430}$\pm$\textbf{0.014} & \textbf{0.472}$\pm$\textbf{0.014} & \multicolumn{1}{c|}{\textbf{0.423}$\pm$\textbf{0.012}} & \textbf{0.392}$\pm$\textbf{0.012} & \textbf{0.352}$\pm$\textbf{0.014} & \textbf{0.325}$\pm$\textbf{0.010} \\ \hline\end{tabular}
}
\end{table*}

\subsection{Eigen-Projection and Concentration Entropy Analysis on Additional Datasets}
\label{sec:sup_projection}
\begin{figure}[!ht]
\centering
\includegraphics[width=0.95\textwidth]{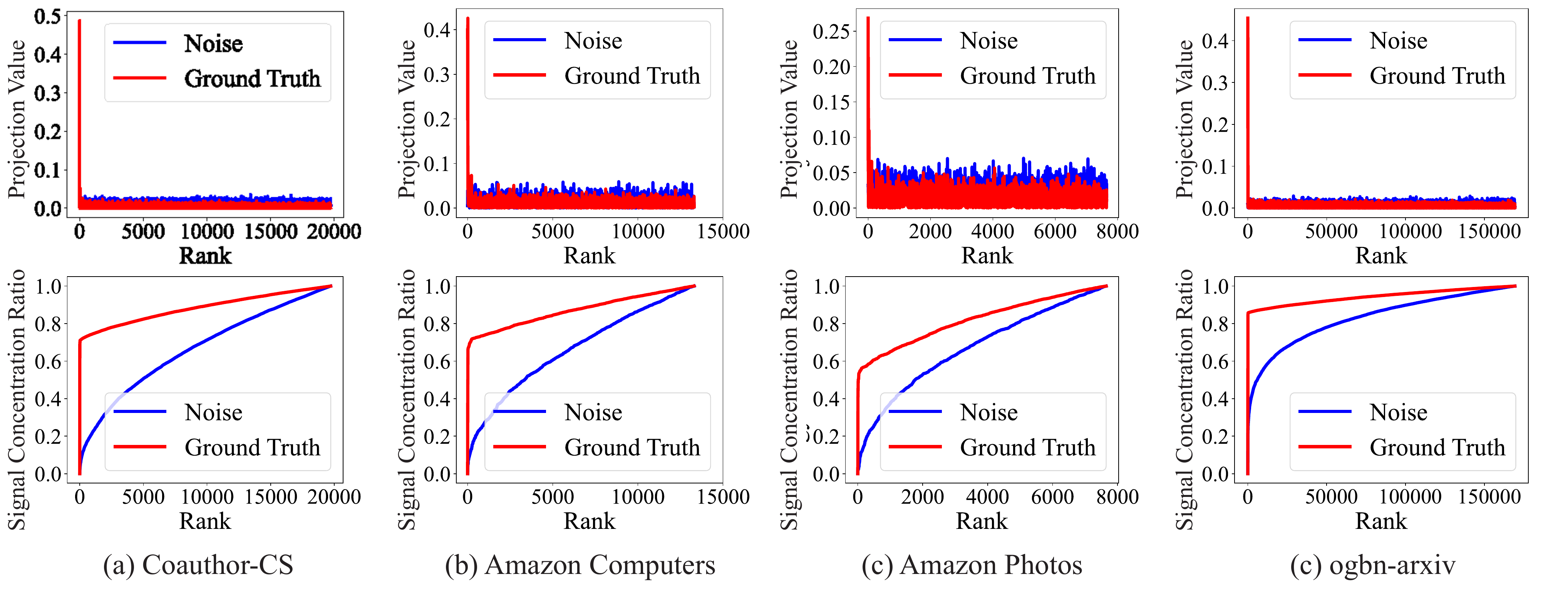}
\caption{Eigen-projection (first row) and energy concentration (second row) on Coauthor-CS, Amazon-Computers, Amazon-Photos, and ogbn-arxiv.
By the rank of $0.2 \min\set{N,d}$, the concentration entropy on Coauthor-CS, Amazon-Computers, Amazon-Photos, and ogbn-arxiv are $0.779$, $0.809$, $0.752$, and $0.787$.}
\label{fig:eigen-projection-more}
\end{figure}

Figure~\ref{fig:eigen-projection-more} presents the eigen-projection visualizations and corresponding signal concentration ratios for Coauthor-CS, Amazon-Computers, Amazon-Photos, and ogbn-arxiv.

\subsection{Evaluation on Heterophilic Graphs}
\label{sec:heterophlic}
\begin{figure}[!ht]
\centering
\includegraphics[width=0.6\textwidth]{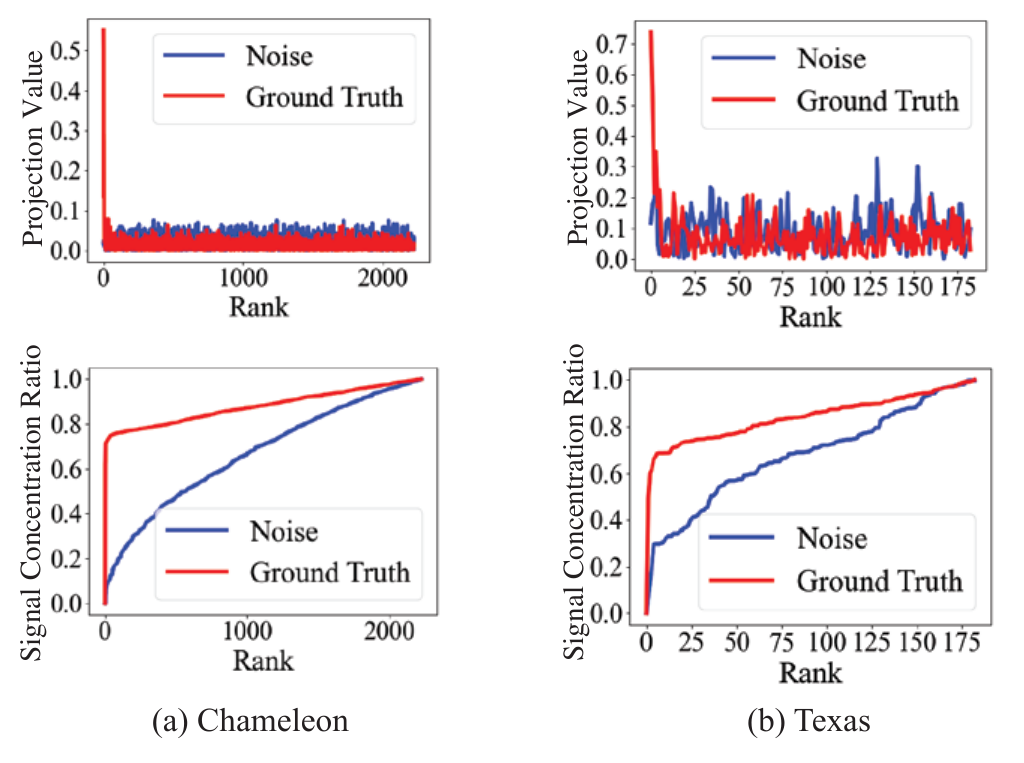}
\caption{Eigen-projection (first row) and signal concentration ratio (second row) on Chameleon and Texas.
The study in this figure is performed for asymmetric label noise with a noise level of $60\%$. By the rank of $0.2 \min\set{N,d}$, the concentration entropy on Chameleon and Texas are $0.762$ and $0.725$.}
\label{fig:eigen-projection-heter}
\end{figure}

In this section, we examine the effectiveness of GCL-LRR for semi-supervised node classification on two commonly used heterophilic graph datasets: Texas and Chameleon~\citep{PeiWCLY20}. We begin by analyzing the Low Frequency Property (LFP) on these datasets using eigen-projection visualizations and signal concentration ratios, as depicted in Figure~\ref{fig:eigen-projection-heter}. The results indicate that the LFP is also present in heterophilic graphs, akin to its presence in homophilic datasets. This analysis is conducted under asymmetric label noise with a corruption level of $60\%$. Using a rank parameter of $0.2 \min\set{N,d}$, the resulting concentration entropy values are $0.762$ for Chameleon and $0.725$ for Texas.

Subsequently, we conduct semi-supervised node classification experiments on Texas and Chameleon following the experimental protocol described in Section~\ref{sec:classification}. We employ TEDGCN~\citep{TEDGCN}, a widely adopted GNN architecture for heterophilic graphs, as the encoder backbone for both GCL-LRR and GCL-LR-Attention. The classification results are reported in Table~\ref{tab:heterophlic}. As shown, both GCL-LRR and GCL-LR-Attention achieve substantial performance improvements over the base heterophilic GNN, demonstrating robustness to various types of noise in these challenging graph settings.

\begin{table*}[!htb]
\large
\centering
\caption{Performance comparison for node classification on Texas and Chameleon with asymmetric label noise, symmetric label noise, and attribute noise.}
\label{tab:heterophlic}
\resizebox{\textwidth}{!}{
\begin{tabular}{|c|c|cccccccccc|}
\hline
\multirow{3}{*}{Dataset}   & \multirow{3}{*}{Methods} & \multicolumn{10}{c|}{Noise Type}                                                                                                                                                  \\ \cmidrule{3-12}
                           &                          & \multicolumn{1}{c|}{0}    & \multicolumn{3}{c|}{40}                                 & \multicolumn{3}{c|}{60}                                 & \multicolumn{3}{c|}{80}            \\ \cmidrule{3-12}
                           &                          & \multicolumn{1}{c|}{-}    & Asymmetric & Symmetric & \multicolumn{1}{c|}{Attribute} & Asymmetric & Symmetric & \multicolumn{1}{c|}{Attribute} & Asymmetric & Symmetric & Attribute \\ \hline
\multirow{3}{*}{Texas}     & TEDGCN                   & \multicolumn{1}{c|}{0.771$\pm$0.025} & 0.525$\pm$0.023 & 0.528$\pm$0.018 & \multicolumn{1}{c|}{0.541$\pm$0.022} & 0.402$\pm$0.016 & 0.418$\pm$0.019 & \multicolumn{1}{c|}{0.445$\pm$0.021} & 0.312$\pm$0.015 & 0.328$\pm$0.017 & 0.341$\pm$0.020 \\
                           & GCL-LRR                   & \multicolumn{1}{c|}{\underline{0.780$\pm$0.013}} & \underline{0.547$\pm$0.019} & \underline{0.557$\pm$0.016} & \multicolumn{1}{c|}{\underline{0.568$\pm$0.017}} & \underline{0.438$\pm$0.015} & \underline{0.444$\pm$0.017} & \multicolumn{1}{c|}{\underline{0.463$\pm$0.018}} & \underline{0.336$\pm$0.012} & \underline{0.353$\pm$0.014} & \underline{0.365$\pm$0.016} \\
                           & GCL-LR-Attention               & \multicolumn{1}{c|}{\textbf{0.785}$\pm$\textbf{0.018}} & \textbf{0.556}$\pm$\textbf{0.016} & \textbf{0.563}$\pm$\textbf{0.013} & \multicolumn{1}{c|}{\textbf{0.576}$\pm$\textbf{0.015}} & \textbf{0.451}$\pm$\textbf{0.012} & \textbf{0.452}$\pm$\textbf{0.014} & \multicolumn{1}{c|}{\textbf{0.472}$\pm$\textbf{0.016}} & \textbf{0.338}$\pm$\textbf{0.010} & \textbf{0.367}$\pm$\textbf{0.012} & \textbf{0.372}$\pm$\textbf{0.014} \\ \hline
\multirow{3}{*}{Chameleon} & TEDGCN                   & \multicolumn{1}{c|}{0.569$\pm$0.009} & 0.382$\pm$0.021 & 0.401$\pm$0.018 & \multicolumn{1}{c|}{0.425$\pm$0.020} & 0.298$\pm$0.017 & 0.315$\pm$0.019 & \multicolumn{1}{c|}{0.328$\pm$0.022} & 0.225$\pm$0.016 & 0.241$\pm$0.018 & 0.254$\pm$0.021 \\
                           & GCL-LRR                   & \multicolumn{1}{c|}{\underline{0.584$\pm$0.011}} & \underline{0.407$\pm$0.019} & \underline{0.436$\pm$0.015} & \multicolumn{1}{c|}{\underline{0.447$\pm$0.018}} & \underline{0.332$\pm$0.015} & \underline{0.342$\pm$0.016} & \multicolumn{1}{c|}{\underline{0.356$\pm$0.018}} & \underline{0.251$\pm$0.013} & \underline{0.269$\pm$0.015} & \underline{0.283$\pm$0.017} \\
                           & GCL-LR-Attention               & \multicolumn{1}{c|}{\textbf{0.585}$\pm$\textbf{0.008}} & \textbf{0.412}$\pm$\textbf{0.016} & \textbf{0.444}$\pm$\textbf{0.013} & \multicolumn{1}{c|}{\textbf{0.452}$\pm$\textbf{0.014}} & \textbf{0.341}$\pm$\textbf{0.011} & \textbf{0.352}$\pm$\textbf{0.013} & \multicolumn{1}{c|}{\textbf{0.361}$\pm$\textbf{0.015}} & \textbf{0.262}$\pm$\textbf{0.010} & \textbf{0.282}$\pm$\textbf{0.012} & \textbf{0.290}$\pm$\textbf{0.014} \\ \hline
\end{tabular}

}
\end{table*}

\subsection{Node Classification Results for GCL Methods with Different Types of Classifiers}
\label{sec:trans_classifier}
Existing GCL approaches, including MERIT~\citep{Jin2021MultiScaleCS}, SUGRL~\citep{mo2022simple}, and SFA~\citep{zhang2023spectral}, typically follow a two-stage procedure: they first train a graph encoder using contrastive objectives such as InfoNCE~\citep{Jin2021MultiScaleCS}, and subsequently train a linear classifier in a supervised manner on the resulting node representations. In contrast, GCL-LRR integrates a transductive classifier directly atop the contrastively learned representations, enabling label propagation during training. To ensure a fair comparison, we retrain all baseline GCL methods using the same transductive classifier employed in GCL-LRR, as well as an additional two-layer transductive GCN classifier. The results with different types of classifiers are shown in Table~\ref{table:classifier_study}. It is observed that LRA-GCL still outperforms SOTA GCL methods even when the transductive classifiers are employed.
\begin{table*}[!htbp]
\centering
\caption{Performance comparison for node classification by inductive linear classifier, transductive two-layer GCN classifier, and transductive classifier used in GCL-LRR. The comparisons are performed on Cora.}
\label{table:classifier_study}
\resizebox{\columnwidth}{!}{
\begin{tabular}{|c|cccccccccc|}
\hline
  &  \multicolumn{10}{c|}{Noise Type}    \\  \cmidrule{2-11}
 &  \multicolumn{1}{c|}{0}  &  \multicolumn{3}{c|}{40}  &  \multicolumn{3}{c|}{60}  &  \multicolumn{3}{c|}{80}  \\  \cmidrule{2-11}
 \multirow{-3}{*}{Methods}  &  \multicolumn{1}{c|}{-}  &  Asymmetric  &  Symmetric  &  \multicolumn{1}{c|}{Attribute}  &  Asymmetric  &  Symmetric  &  \multicolumn{1}{c|}{Attribute}  &  Asymmetric  &  Symmetric  &  Attribute  \\  \hline
 SUGRL (original, inductive classifier)                                 &  \multicolumn{1}{c|}{0.834$\pm$0.005}  &  0.564$\pm$0.011  &  0.674$\pm$0.012  &  \multicolumn{1}{c|}{0.675$\pm$0.009}  &  0.468$\pm$0.011  &  0.552$\pm$0.011  &  \multicolumn{1}{c|}{0.452$\pm$0.012}  &  0.280$\pm$0.012  &  0.381$\pm$0.012  &  0.338$\pm$0.014  \\
 SUGRL + transductive GCN                       &  \multicolumn{1}{c|}{0.833$\pm$0.006}  &  0.562$\pm$0.013  &  0.675$\pm$0.015  &  \multicolumn{1}{c|}{0.673$\pm$0.012}  &  0.470$\pm$0.011  &  0.551$\pm$0.011  &  \multicolumn{1}{c|}{0.454$\pm$0.012}  &  0.280$\pm$0.012  &  0.380$\pm$0.012  &  0.340$\pm$0.014  \\
 SUGRL + linear transductive classifier         &  \multicolumn{1}{c|}{0.836$\pm$0.007}  &  0.568$\pm$0.013  &  0.677$\pm$0.010  &  \multicolumn{1}{c|}{0.674$\pm$0.011}  &  0.472$\pm$0.011  &  0.555$\pm$0.011  &  \multicolumn{1}{c|}{0.457$\pm$0.012}  &  0.284$\pm$0.012  &  0.383$\pm$0.012  &  0.341$\pm$0.014  \\
 \hline
 MERIT (original, inductive classifier)                         &  \multicolumn{1}{c|}{0.831$\pm$0.005}  &  0.560$\pm$0.008  &  0.670$\pm$0.008  &  \multicolumn{1}{c|}{0.671$\pm$0.009}  &  0.467$\pm$0.013  &  0.547$\pm$0.013  &  \multicolumn{1}{c|}{0.450$\pm$0.014}  &  0.277$\pm$0.013  &  0.385$\pm$0.013  &  0.335$\pm$0.009  \\
 MERIT + transductive GCN                       &  \multicolumn{1}{c|}{0.831$\pm$0.007}  &  0.562$\pm$0.011  &  0.668$\pm$0.013  &  \multicolumn{1}{c|}{0.672$\pm$0.014}  &  0.466$\pm$0.013  &  0.549$\pm$0.015  &  \multicolumn{1}{c|}{0.451$\pm$0.016}  &  0.276$\pm$0.012  &  0.382$\pm$0.014  &  0.337$\pm$0.013  \\
 MERIT + linear transductive classifier         &  \multicolumn{1}{c|}{0.833$\pm$0.003}  &  0.562$\pm$0.014  &  0.673$\pm$0.012  &  \multicolumn{1}{c|}{0.673$\pm$0.011}  &  0.466$\pm$0.015  &  0.546$\pm$0.016  &  \multicolumn{1}{c|}{0.453$\pm$0.017}  &  0.280$\pm$0.016  &  0.386$\pm$0.011  &  0.336$\pm$0.014  \\
  \hline
 SFA (original, inductive classifier)                                   &  \multicolumn{1}{c|}{0.839$\pm$0.010}  &  0.564$\pm$0.011  &  0.677$\pm$0.013  &  \multicolumn{1}{c|}{0.676$\pm$0.015}  &  0.473$\pm$0.014  &  0.549$\pm$0.014  &  \multicolumn{1}{c|}{0.457$\pm$0.014}  &  0.282$\pm$0.016  &  0.389$\pm$0.013  &  0.344$\pm$0.017  \\
 SFA + transductive GCN                                 &  \multicolumn{1}{c|}{0.837$\pm$0.013}  &  0.565$\pm$0.011  &  0.673$\pm$0.017  &  \multicolumn{1}{c|}{0.673$\pm$0.018}  &  0.474$\pm$0.016  &  0.551$\pm$0.015  &  \multicolumn{1}{c|}{0.453$\pm$0.018}  &  0.277$\pm$0.016  &  0.389$\pm$0.015  &  0.343$\pm$0.019  \\
 SFA + linear transductive classifier           &  \multicolumn{1}{c|}{0.841$\pm$0.015}  &  0.566$\pm$0.013  &  0.678$\pm$0.014  &  \multicolumn{1}{c|}{0.679$\pm$0.014}  &  0.477$\pm$0.015  &  0.552$\pm$0.012  &  \multicolumn{1}{c|}{0.456$\pm$0.016}  &  0.284$\pm$0.017  &  0.391$\pm$0.015  &  0.348$\pm$0.019  \\
  \hline
  GCL-LRR  &  \multicolumn{1}{c|}{\underline{0.858$\pm$0.006}}  &  \underline{0.589$\pm$0.011}  &  \underline{0.713$\pm$0.007}  &  \multicolumn{1}{c|}{\underline{0.695$\pm$0.011}}  & \underline{0.492$\pm$0.011}  &  \underline{0.587$\pm$0.013}  &  \multicolumn{1}{c|}{\underline{0.477$\pm$0.012}}  &  \underline{0.306$\pm$0.012}  &  \underline{0.419$\pm$0.012}  & \underline{0.363$\pm$0.011}  \\
  GCL-LR-Attention & \multicolumn{1}{c|}{\textbf{0.861}$\pm$\textbf{0.006}} & \textbf{0.602}$\pm$\textbf{0.011} & \textbf{0.724}$\pm$\textbf{0.007} & \multicolumn{1}{c|}{\textbf{0.708}$\pm$\textbf{0.011}} & \textbf{0.510}$\pm$\textbf{0.011} & \textbf{0.605}$\pm$\textbf{0.013} & \multicolumn{1}{c|}{\textbf{0.492}$\pm$\textbf{0.012}} & \textbf{0.329}$\pm$\textbf{0.012} & \textbf{0.436}$\pm$\textbf{0.012} & \textbf{0.382}$\pm$\textbf{0.011} \\ \hline
\end{tabular}
}
\end{table*}

\subsection{Statistical Significance Analysis}
\label{sec:t-test}
In this section, we compute the p-values from paired t-tests comparing the performance of GCL-LRR and GCL-LR-Attention against the strongest baseline methods to assess the statistical significance of the observed improvements. As shown in Table~\ref{table:p-value}, the p-values for both GCL-LRR and GCL-LR-Attention remain consistently below the threshold of $0.05$ across all datasets and noise conditions, thereby confirming that the performance gains over the top baselines are statistically significant.

\begin{table*}[!htb]
\large
\centering
\caption{P-values of the t-tests for GCL-LRR and GCL-LR-Attention against the top baseline methods under each noise setting on all the benchmark datasets.}
\label{table:p-value}
\resizebox{\textwidth}{!}{
\begin{tabular}{|c|c|ccccccccc|}
\hline
\multirow{3}{*}{Datasets}         & \multirow{3}{*}{Methods} & \multicolumn{9}{c|}{Noise Type}                                                                                                                        \\ \cmidrule{3-11}
                                  &                          & \multicolumn{3}{c|}{40}                                 & \multicolumn{3}{c|}{60}                                 & \multicolumn{3}{c|}{80}            \\ \cmidrule{3-11}
                                  &                          & Asymmetric & Symmetric & \multicolumn{1}{c|}{Attribute} & Asymmetric & Symmetric & \multicolumn{1}{c|}{Attribute} & Asymmetric & Symmetric & Attribute \\ \hline
\multirow{2}{*}{Cora}             & GCL-LRR                   & 0.038      & 0.024     & \multicolumn{1}{c|}{0.021}     & 0.035      & 0.021     & \multicolumn{1}{c|}{0.041}     & 0.025      & 0.028     & 0.031     \\
                                  & GCL-LR-Attention                  & 0.027      & 0.022     & \multicolumn{1}{c|}{0.018}     & 0.018      & 0.038     & \multicolumn{1}{c|}{0.035}     & 0.031      & 0.023     & 0.030     \\ \hline
\multirow{2}{*}{Citeseer}         & GCL-LRR                   & 0.043      & 0.035     & \multicolumn{1}{c|}{0.022}     & 0.021      & 0.030     & \multicolumn{1}{c|}{0.041}     & 0.027      & 0.024     & 0.022     \\
                                  & GCL-LR-Attention                  & 0.037      & 0.038     & \multicolumn{1}{c|}{0.019}     & 0.027      & 0.025     & \multicolumn{1}{c|}{0.040}     & 0.035      & 0.037     & 0.030     \\ \hline
\multirow{2}{*}{PubMed}           & GCL-LRR                   & 0.028      & 0.043     & \multicolumn{1}{c|}{0.030}     & 0.026      & 0.027     & \multicolumn{1}{c|}{0.043}     & 0.036      & 0.040     & 0.042     \\
                                  & GCL-LR-Attention                  & 0.025      & 0.030     & \multicolumn{1}{c|}{0.026}     & 0.023      & 0.024     & \multicolumn{1}{c|}{0.041}     & 0.033      & 0.035     & 0.037     \\ \hline
\multirow{2}{*}{Coauthor-CS}      & GCL-LRR                   & 0.041      & 0.032     & \multicolumn{1}{c|}{0.036}     & 0.043      & 0.044     & \multicolumn{1}{c|}{0.040}     & 0.027      & 0.037     & 0.042     \\
                                  & GCL-LR-Attention                  & 0.036      & 0.030     & \multicolumn{1}{c|}{0.034}     & 0.041      & 0.042     & \multicolumn{1}{c|}{0.039}     & 0.025      & 0.033     & 0.036     \\ \hline
\multirow{2}{*}{ogbn-arxiv}       & GCL-LRR                   & 0.040      & 0.032     & \multicolumn{1}{c|}{0.036}     & 0.043      & 0.044     & \multicolumn{1}{c|}{0.040}     & 0.027      & 0.037     & 0.042     \\
                                  & GCL-LR-Attention                  & 0.036      & 0.030     & \multicolumn{1}{c|}{0.034}     & 0.041      & 0.042     & \multicolumn{1}{c|}{0.039}     & 0.025      & 0.033     & 0.036     \\ \hline
\multirow{2}{*}{Wiki-CS}          & GCL-LRR                   & 0.044      & 0.035     & \multicolumn{1}{c|}{0.033}     & 0.028      & 0.034     & \multicolumn{1}{c|}{0.041}     & 0.036      & 0.040     & 0.042     \\
                                  & GCL-LR-Attention                  & 0.040      & 0.036     & \multicolumn{1}{c|}{0.031}     & 0.026      & 0.029     & \multicolumn{1}{c|}{0.039}     & 0.034      & 0.038     & 0.040     \\ \hline
\multirow{2}{*}{Amazon-Computers} & GCL-LRR                   & 0.033      & 0.043     & \multicolumn{1}{c|}{0.034}     & 0.031      & 0.034     & \multicolumn{1}{c|}{0.041}     & 0.036      & 0.040     & 0.042     \\
                                  & GCL-LR-Attention                  & 0.031      & 0.038     & \multicolumn{1}{c|}{0.030}     & 0.028      & 0.030     & \multicolumn{1}{c|}{0.038}     & 0.034      & 0.037     & 0.040     \\ \hline
\multirow{2}{*}{Amazon-Photos}    & GCL-LRR                   & 0.040      & 0.041     & \multicolumn{1}{c|}{0.038}     & 0.043      & 0.044     & \multicolumn{1}{c|}{0.040}     & 0.027      & 0.037     & 0.042     \\
                                  & GCL-LR-Attention                  & 0.037      & 0.039     & \multicolumn{1}{c|}{0.036}     & 0.041      & 0.042     & \multicolumn{1}{c|}{0.039}     & 0.025      & 0.033     & 0.036     \\ \hline
\multirow{2}{*}{Texas}            & GCL-LRR                   & 0.038      & 0.024     & \multicolumn{1}{c|}{0.021}     & 0.035      & 0.021     & \multicolumn{1}{c|}{0.041}     & 0.025      & 0.028     & 0.031     \\
                                  & GCL-LR-Attention                  & 0.027      & 0.022     & \multicolumn{1}{c|}{0.018}     & 0.018      & 0.038     & \multicolumn{1}{c|}{0.035}     & 0.031      & 0.023     & 0.030     \\ \hline
\multirow{2}{*}{Chameleon}        & GCL-LRR                   & 0.044      & 0.035     & \multicolumn{1}{c|}{0.033}     & 0.028      & 0.034     & \multicolumn{1}{c|}{0.041}     & 0.036      & 0.040     & 0.042     \\
                                  & GCL-LR-Attention                  & 0.040      & 0.036     & \multicolumn{1}{c|}{0.031}     & 0.026      & 0.029     & \multicolumn{1}{c|}{0.039}     & 0.034      & 0.038     & 0.040     \\ \hline
\end{tabular}
}
\end{table*}
\begin{table*}[!htpb]
\small
\centering
\caption{Sensitivity analysis on the weighting parameter $\tau$ for the TNN. The study is performed using GCL-LR-Attention on the Coauthor-CS dataset for semi-supervised node classification under $60\%$ asymmetric label noise.}
\label{tab:sensitivity}
\resizebox{0.775\textwidth}{!}{
\begin{tabular}{|c|ccccccccc|}
\hline
$\tau$ & 0.1   & 0.2   & 0.3   & 0.4   & 0.5   & 0.6   & 0.7   & 0.8   & 0.9  \\
\hline
Accuracy            & 0.588 & 0.590 & 0.593 & 0.591 & 0.594 & 0.590 & 0.591 & 0.590 & 0.589\\
\hline

\end{tabular}
        }
\end{table*}

\subsection{Sensitivity Analysis on the Hyperparameters}
\label{sec:sensitivity}
We perform a sensitivity analysis on the weighting parameter $\tau$ associated with the TNN $\norm{\bK}{r_0}$ in Equation~\ref{eq:loss-GCL-LR-overall}. This analysis is conducted using GCL-LR-Attention on the Coauthor-CS dataset under the setting of semi-supervised node classification with $60\%$ asymmetric label noise. We vary $\tau$ over the set $\{0.1, 0.2, 0.3, 0.4, 0.5, 0.6, 0.7, 0.8, 0.9\}$, and the corresponding classification accuracies are reported in Table~\ref{tab:sensitivity}. The highest accuracy is achieved at $\tau = 0.5$, though GCL-LR-Attention maintains stable and competitive performance across the entire range. Even in the least favorable case, with $\tau = 0.1$, the accuracy declines by only $0.6\%$, highlighting the robustness of the model to variations in $\tau$.

\section{Conclusions}
This paper first introduces a novel graph contrastive learning encoder, Graph Contrastive Learning with Low-Rank Regularization (GCL-LRR), designed to enhance the robustness in node representation learning. GCL-LRR generates low-rank features, drawing motivation from the low-frequency characteristics commonly observed in real-world graph datasets and the tight generalization bound for transductive learning. The GCL-LRR encoder is trained under a prototypical GCL framework, incorporating the TNN as a regularization term. We assess the effectiveness of GCL-LRR through extensive comparisons with existing methods on semi-supervised and transductive node classification tasks, where the input graphs are perturbed by either label noise or attribute noise.
To further strengthen its performance, we extend GCL-LRR with a novel low-rank attention mechanism, leading to the GCL-LR-Attention model. GCL-LR-Attention further reduces the kernel complexity of GCL-LRR, thereby improving the generalization of GCL-LRR. Empirical results across a wide range of benchmarks confirm that both GCL-LRR and GCL-LR-Attention achieve superior robustness and outperform state-of-the-art approaches in learning effective node representations for noisy node classification tasks.

\backmatter

\section{Theoretical Results}
\label{sec:transductive-theory}

We present the proof of Theorem~\ref{theorem:optimization-linear-kernel} in this section.

\begin{proof}[\textup{\bf Proof of Theorem~\ref{theorem:optimization-linear-kernel}}]
Define $\bN \defeq \bY - \tilde \bY \in \RR^N$ as the label noise.
It can be verified that at the $t$-th iteration of gradient descent for $t \ge 1$, we have
\bal\label{eq:optimization-linear-kernel-seg1}
\bW^{(t)} &= \bW^{(t-1)} - \eta \bth{\bH}_{\cL}^{\top}   \bth{\bH \bW^{(t-1)} - \bY}_{\cL} \nonumber \\
&= \bW^{(t-1)} - \eta \bth{\bH}_{\cL}^{\top}   \bth{\bH \bW^{(t-1)} -
\tilde \bY}_{\cL} + \eta \bth{\bH}_{\cL}^{\top} \bth{\bN}_{\cL}.
\eal
It follows by (\ref{eq:optimization-linear-kernel-seg1}) that
\bal\label{eq:optimization-linear-kernel-seg2}
\bth{\bH}_{\cL} \bW^{(t)} =  \bth{\bH}_{\cL} \bW^{(t-1)} - \eta
{\bK}_{\cL,\cL} \bth{\bH \bW^{(t-1)} - \tilde \bY}_{\cL} +  \eta
\bth{\bK}_{\cL,\cL}\bth{\bN}_{\cL},
\eal
where $\bK_{\cL,\cL}
\defeq \bth{\bH}_{\cL}
\bth{\bH}_{\cL}^{\top} \in \RR^{m \times m}$. With $\bF(\bW,t) = \bH \bW^{(t)}$,  it follows by (\ref{eq:optimization-linear-kernel-seg2}) that
\bals
\bth{\bF(\bW,t) - \tilde \bY}_{\cL} = \pth{\bI_ m- \eta \bth{\bK}_{\cL,\cL} } \bth{\bF(\bW,t-1) - \tilde \bY}_{\cL} + \eta
\bth{\bK}_{\cL,\cL}\bth{\bN}_{\cL}.
\eals
It follows from the above equality and the recursion that
\bal\label{eq:optimization-linear-kernel-full-loss}
\bth{\bF(\bW,t) - \tilde \bY}_{\cL} = - \pth{\bI_ m- \eta \bth{\bK}_{\cL,\cL} }^t \bth{\tilde \bY}_{\cL}
+  \eta \bth{\bK}_{\cL,\cL} \sum_{t'=0}^{t-1} \pth{\bI_ m- \eta \bth{\bK}_{\cL,\cL} }^{t'}\bth{\bN}_{\cL}
\eal
We apply \citep[Corollary 3.7]{TLRC} to obtain the following bound for the test loss $\frac 1u \fnorm{\bth{\bF(\bW,t) - \tilde \bY}_{\cU}}^2$:
\bal\label{eq:optimization-linear-kernel-test-loss-1}
\frac 1u \fnorm{\bth{\bF(\bW,t) - \tilde \bY}_{\cU}}^2 &\le \frac {c_0}m \fnorm{\bth{\bF(\bW,t) - \tilde \bY}_{\cL}}^2  + c_0 \min_{0 \le Q \le N} r(u,m,Q)+ \frac{c_0x}{u},
\eal
with
\bals
r(u,m,Q) \defeq Q \pth{\frac{1}{u} + \frac{1}{m}} + \pth{ \sqrt{\frac{\sum\limits_{q = Q+1}^N\hat \lambda_q}{u}}
+ \sqrt{\frac{\sum\limits_{q = Q+1}^N\hat \lambda_q}{m}}},
\eals
where $c_0$ is a positive constant depending on $\bU$, $\set{\hat \lambda_i}_{i=1}^r$, and $\tau_0$ with $\tau_0^2 = \max_{i \in [N]} \bK_{ii}$.

It follows from (\ref{eq:optimization-linear-kernel-full-loss}) and
 (\ref{eq:optimization-linear-kernel-test-loss-1}) that for every $r_0 \in [N]$, we have
\bal\label{eq:optimization-linear-kernel-seg3}
&\frac 1u \fnorm{\bth{\bF(\bW,t) - \tilde \bY}_{\cU}}^2 \le \frac {c_0}m \fnorm{\pth{\bI_ m- \eta \bth{\bK}_{\cL,\cL} }^t \bth{\tilde \bY}_{\cL}}^2  \nonumber \\
&\hspace{1.3in}+ c_0 r_0 \pth{\frac{1}{u} + \frac{1}{m}} + c_0 \pth{ \sqrt{\frac{\sum\limits_{q = r_0+1}^N\hat \lambda_q}{u}}
+ \sqrt{\frac{\sum\limits_{q = r_0+1}^N\hat \lambda_q}{m}}} + \frac{c_0x}{u} \nonumber \\
&\stackrel{\circled{1}}{\le} \frac {2c_0}m \fnorm{\pth{\bI_ m- \eta \bth{\bK}_{\cL,\cL} }^t \bth{\tilde \bY}_{\cL}}^2  +\frac {2c_0}m  \fnorm{\eta \bth{\bK}_{\cL,\cL} \sum_{t'=0}^{t-1} \pth{\bI_ m- \eta \bth{\bK}_{\cL,\cL} }^{t'}\bth{\bN}_{\cL}}^2 \nonumber \\
&\phantom{=}+ c_0 r_0 \pth{\frac{1}{u} + \frac{1}{m}} + c_0 \sqrt{\norm{\bK}{r_0}} \pth{ \sqrt{\frac{1}{u}}
+ \sqrt{\frac{1}{m}}} + \frac{c_0x}{u},
\eal
where $\circled{1}$ follows from the Cauchy-Schwarz inequality, (\ref{eq:optimization-linear-kernel-full-loss}), and $\sum_{q = r_0+1}^N \hat \lambda_q = \norm{\bK}{r_0}$.
(\ref{eq:optimization-linear-kernel-test-loss}) then follows directly from
(\ref{eq:optimization-linear-kernel-seg3}).
\end{proof}

\bibliography{sn-bibliography}








\end{document}